\documentclass[preprint,authoryear]{elsarticle}

\usepackage{graphicx}
\usepackage{amssymb}

\usepackage[USenglish]{babel}

\usepackage{booktabs}

\usepackage[T1]{fontenc}
\usepackage{times}
\usepackage{graphicx}
\usepackage{amsmath}
\usepackage{amsfonts}
\usepackage{wasysym}
\usepackage{url}
\usepackage{makecell}
\usepackage{rotating}

\usepackage[boxed,vlined,linesnumbered,ruled]{algorithm2e}
\let\oldnl\nl
\newcommand{\nonl}{\renewcommand{\nl}{\let\nl\oldnl}}

\usepackage{color}
\usepackage{ucs}
\usepackage{subfigure}
\usepackage{latexsym}
\usepackage{multirow}
\usepackage[autolanguage]{numprint}
\usepackage{array}

\usepackage{sidecap} 

\usepackage{times}
\usepackage{xcolor,colortbl}
\definecolor{green}{rgb}{0.1,0.1,0.1}

\usepackage[utf8x]{inputenc}

\definecolor{darkblue}{HTML}{000099}
\usepackage[colorlinks=true,linkcolor=black,citecolor=darkblue]{hyperref}

\DeclareMathOperator{\fst}{fst}
\DeclareMathOperator{\prv}{prv}
\DeclareMathOperator{\nxt}{nxt}
\DeclareMathOperator{\w}{w}
\DeclareMathOperator{\s}{s}
\DeclareMathOperator{\departureTime}{departureTime}
\DeclareMathOperator{\discardedTime}{discardedTime}
\DeclareMathOperator{\load}{load}

\usepackage{enumerate}

\makeindex

\journal{Transportation Research: Part E}

\begin{document}

\begin{frontmatter}


\title{The Static and Stochastic VRPTW\\with both random Customers and Reveal Times:\\algorithms and recourse strategies}


\author[UCLouvain,INSA]{Michael Saint-Guillain}
\ead{michael.saint@uclouvain.be}
\author[INSA,Lyon]{Christine Solnon}
\ead{christine.solnon@insa-lyon.fr}
\author[UCLouvain]{Yves Deville}

\address[UCLouvain]{Université catholique de Louvain, ICTEAM Institute, B-1348 Louvain-la-Neuve, Belgium}
\address[INSA]{INSA-Lyon, LIRIS Laboratory, F-69621, France}
\address[Lyon]{Université de Lyon, CNRS, France}

\begin{abstract}
Unlike its deterministic counterpart, static and stochastic vehicle routing problems (SS-VRP) aim at modeling and solving real-life operational problems by considering uncertainty on data. 
We consider the SS-VRPTW-CR introduced in \cite{Saint-Guillain2017}.
Like the SS-VRP introduced by \cite{Bertsimas1992}, we search for optimal first stage routes for a fleet of vehicles to handle a set of stochastic customer demands, {\em i.e.}, demands are uncertain and we only know their probabilities.
In addition to capacity constraints, customer demands are also constrained by time windows. 
Unlike all SS-VRP variants, the SS-VRPTW-CR does not make any assumption on the time at which a stochastic demand is revealed, {\em i.e.}, the reveal time is stochastic as well. 
To handle this new problem, we introduce waiting locations: Each vehicle is assigned a sequence of waiting locations from which it may serve some associated demands, and the objective is to minimize the expected number of demands that cannot be satisfied in time.
In this paper, we propose two new recourse strategies for the SS-VRPTW-CR, together with their closed-form expressions for efficiently computing their expectations: The first one allows us to take vehicle capacities into account; The second one allows us to optimize routes by avoiding some useless trips.
We propose two algorithms for searching for routes with optimal expected costs: The first one is  an extended branch-and-cut algorithm, based on a stochastic integer formulation, and the second one is a local search based heuristic method.
We also introduce a new public benchmark for the SS-VRPTW-CR, based on real-world data coming from the city of Lyon. We evaluate our two algorithms on this benchmark and empirically demonstrate the expected superiority of the SS-VRPTW-CR anticipative actions over a basic "wait-and-serve" policy.
\end{abstract}


\end{frontmatter}


\section{Introduction \label{sec:Introduction}}
The Vehicle Routing Problem (VRP) aims at modeling and solving a real-life common operational problem, in which a set of customers must be visited using a fleet of vehicles. 
Each customer comes with a certain demand. 
In the VRP with Time Windows (VRPTW), each customer must be visited within a given time window. 
A feasible solution of the VRPTW is a set of vehicle routes, such that every customer is visited exactly once during its time window and that sum of the demands along each route does not exceed the corresponding vehicle's capacity. 
The objective is then to find an optimal feasible solution, where optimality is usually defined in terms of travel distances. 
 
The classical deterministic VRP(TW) assumes that all input data are known with certainty before the computation of the solution. 
However, in real-world applications some input data may be uncertain when computing a solution. 
For instance, only a subset of the customer demands may be known before online execution. 
Missing demands hence arrive in a dynamic fashion, while vehicles are on their route. 
In such a context, a solution should contain operational decisions that deal with current known demands, but should also anticipate potential unknown demands. 
Albeit uncertainty may be considered for various input data of the VRP ({\em e.g.}, travel times), we focus on situations where the customer data are unknown a priori, and we assume that we have some probabilistic knowledge on missing data ({\em e.g.}, probability distributions computed from historical data).
This probabilistic knowledge is used to compute a first-stage solution which is adapted online when random variables are realized.

Two different kinds of adaptations may be considered: {\em Dynamic and Stochastic VRP(TW)} (DS-VRP(TW)) and {\em Static and Stochastic VRP(TW)} (SS-VRP(TW)).
In the DS-VRP(TW), the solution is re-optimized at each time-step, and this re-optimization involves solving an $\cal NP$-hard problem. Therefore, it is usually approximated with meta-heuristics as proposed, for example, in \cite{Ichoua2006exploiting,Bent2007,Saint-Guillain2015}. 

\begin{figure}
\centering
\includegraphics[width=1.0\textwidth]{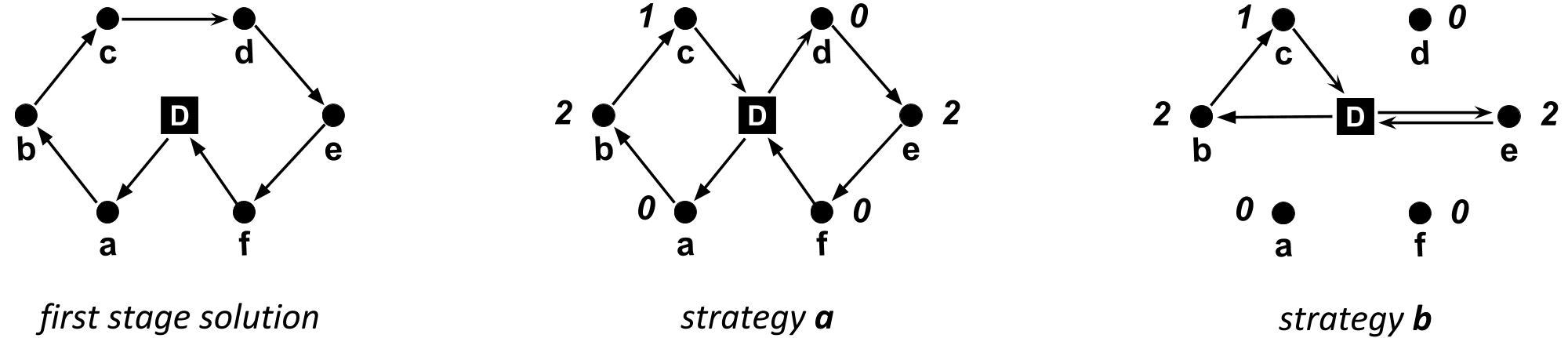}
\caption{
Two examples of recourse strategies taken from \cite{Bertsimas1992}, designed for the SS-VRP with stochastic customers and demands. 
The vehicle has a capacity of 3. 
Based on stochastic information, the first stage solution (\emph{left}) states the {\em a priori} sequence of customer visits.
When applying strategy {\bf a} to adapt this solution (\emph{middle}), we add a return to the depot {\bf D} between customers {\bf c} and {\bf d} to unload the vehicle.
When applying strategy {\bf b} (\emph{right}), we also skip customers {\bf a}, {\bf d} and {\bf f} as their demands are null.
}
\label{fig:example_bertsimas}
\end{figure}

In the SS-VRP(TW), no expensive reoptimization is allowed during online execution. 
When unknown information is revealed, the first stage solution is adapted online by applying a predefined recourse strategy whose time complexity is polynomial.
In this case, the goal is to find a first stage solution that minimizes its total cost plus the \emph{expected} extra cost caused by the recourse strategy. 
For example, in \cite{Bertsimas1992}, the first stage solution is a set of vehicle tours which is computed offline with respect to probability distributions of customer demands. Real customer demands are revealed online, and two different recourse strategies are proposed, as illustrated in Fig. \ref{fig:example_bertsimas}: In the first one (strategy \textbf{a}), each demand is assumed to be known when the vehicle arrives at the customer place, and if it is larger than or equal to the remaining capacity of the vehicle, then the first stage solution is adapted by adding a round trip to the depot to unload the vehicle; In the second recourse strategy (strategy \textbf{b}), each demand is assumed to be known when leaving the previous customer and the recourse strategy is refined to skip customers with null demands. 
More recently, a preventive restocking strategy for the SS-VRP with random demands has been proposed by \cite{Yang2000}. 
\cite{Biesinger2016} later introduced a variant for the Generalized SS-VRP with random demands.


In recent review \cite{Gendreau2016}, the authors argue for new recourse strategies: With the increasing use of ICT, customer demand (and by extension presence) information is likely to be revealed on a very frequent basis. 
In this context, the chronological order in which this information is transmitted no longer matches the planned sequences of customers on the vehicle routes.
In particular, the authors consider as paradoxical the fact that the existing literature on SS-VRPs with random Customers (SS-VRP-C) assumes full knowledge on the presence of customers at the beginning of the operational period.

In this paper, we focus on the SS-VRPTW with both random Customers and Reveal times (SS-VRPTW-CR) recently introduced in \cite{Saint-Guillain2017}. The SS-VRPTW-CR does not make strong assumptions on the moment at which customer requests are revealed during the operations, contrary to most existing work that assume that customer requests are known either when arriving at the customer place, or when leaving the previous customer.
To handle uncertainty on reveal times, waiting locations are introduced when computing first-stage solutions: The routes computed offline visit waiting locations and a waiting time is associated with each waiting location. 
When a customer request is revealed, it is either accepted (if it is possible to serve it) or rejected. 
The recourse strategy then adapts routes in such a way that all accepted requests are guaranteed to eventually be served. 
The goal is to compute the first-stage solution that minimizes the expected number of rejected requests.

An example of practical application of the SS-VRPTW-CR is an on-demand health care service for elderly or disabled people. 
Health care services are provided directly at home by mobile medical units.
Every registered person may request a health care support at any moment of the day with the guarantee to be satisfied within a given time window.
From historical data, we know the probability that a request appears for each customer and each time unit.
Given this stochastic knowledge, we compute a first-stage solution.
When a request appears (online), the recourse strategy is used to decide whether the request is accepted or rejected and to adapt medical unit routes.
When a request is rejected, the system must rely on an external service provider in order to satisfy it.  
Therefore, the goal is to minimize the expected number of rejected requests.


\paragraph{Contributions}
Up to our knowledge, previous static and stochastic VRP's studies all assume that requests are revealed at the beginning of the day, and all fail at capturing the following property:
Besides the stochasticity on request presence, the moment at which a request is received is stochastic as well. 
The SS-VRPTW-CR recently introduced in \cite{Saint-Guillain2017} is the first one that actually captures this property. However, the recourse strategy proposed in this paper does not take capacity constraints into account. Hence, a first contribution is to introduce a new recourse strategy that handles these constraints. We also introduce an improved recourse strategy that optimizes routes by skipping some useless parts. We introduce closed-form expressions to efficiently compute expected costs for these two new recourse strategies.
We also propose a stochastic integer programming formulation and an extended branch-and-cut algorithm for these recourse strategies. 

Another contribution is to introduce a new public benchmark for the SS-VRPTW-CR, based on real-word data coming from the city of Lyon. By comparing with a basic (yet realistic) wait-and-serve policy which does not exploit stochastic knowledge, computational experiments on this benchmark show that the models we propose behave better in average. 
While the exact algorithm fails at scaling to realistic problem sizes, we show that the heuristic local search approach proposed in \cite{Saint-Guillain2017} not only gives near-optimal solutions on small instances, but also leads to promising results on bigger ones.
Improved variants of the heuristic method are then described and their efficiency demonstrated on the bigger instances.
Experiments indicate that using a SS-VRPTW-CR model is particularly beneficial as the number of vehicles increases and when the time windows impose a high level of responsiveness.
Eventually, all the experiments show that allowing the vehicles to wait directly at potential customer locations lead to better expected results than using separated relocation vertices.

\paragraph{Organization}
In section \ref{sec:related-work}, we review the existing studies on VRPs that imply stochastic customers, and clearly position the SS-VRPTW-CR with respect to them.
Section \ref{sec:Problem-description} formally defines the general SS-VRPTW-CR. 
Section \ref{sec:R1_infinite_capacity} describes the recourse strategy already introduced in \cite{Saint-Guillain2017}, which applies when there is no constraint on the vehicle capacities.
Sections \ref{sec:R_Q} and \ref{sec:R_Q_plus} introduce two new recourse strategies to deal with limited vehicle capacities and more clever vehicle operations.
Section \ref{sec:exact} introduces a stochastic integer programming formulation and a branch-and-cut based solving method for solving the problem to optimality. 
Section \ref{sec:local-search} describes the heuristic algorithm of \cite{Saint-Guillain2017} to efficiently find solutions of good quality.
Experimental results are analyzed in section \ref{sec:experimentations}. 
Further research directions are finally discussed in section \ref{sec:conclusions}.

\section{Related work \label{sec:related-work}}

By definition, the SS-VRPTW-CR is a static problem. 
In this section we hence do not consider dynamic VRPs and rather focus on existing studies that have been carried on \emph{static and stochastic VRPs}. Specific literature reviews on the SS-VRP may be found in 
\cite{gendreau1996stochastic}, 
\cite{bertsimas1996new},
\cite{cordeau2007vehicle}, 
\cite{Campbell2008g} and recently in
\cite{toth2014vehicle}, \cite{Berhan2014}, \cite{Kovacs2014} and \cite{Gendreau2016}.
According to \cite{pillac2013review}, the most studied cases in SS-VRPs are: 
\begin{itemize}
\item Stochastic customers (SS-VRP-C), where customer presences are described by random variables; 
\item Stochastic demands (SS-VRP-D), where all customers are present but their demands are random variables (see for instance 
\citeauthor{dror1989vehicle} (\citeyear{dror1989vehicle}, \citeyear{dror1993vehicle}),  
\cite{Laporte2002}, 
\cite{Morales2006}, 
\cite{Christiansen2007}, 
\citeauthor{Mendoza2010} (\citeyear{Mendoza2010}, 
\citeyear{Mendoza2011}),  
\citeauthor{Secomandi2000} (\citeyear{Secomandi2000}, \citeyear{Secomandi2009}) and
\cite{Gauvin2014});
\item Stochastic times (SS-VRP-T), where either travel and/or service times are random variables (see for instance 
\cite{laporte1992vehicle}, 
\cite{kenyon2003stochastic}, 
\cite{verweij2003sample} and 
\cite{li2010vehicle}). 
\end{itemize}
Since the SS-VRPTW-CR belongs to the first category, we focus this review on customers uncertainty only.



The Traveling Salesman Problem (TSP) is a special case of the VRP with only one uncapacitated vehicle.
The first study on static and stochastic vehicle routing is due to \cite{bartholdi1983minimal}, who considered {\em a priori} solutions to daily food delivery. 
\cite{jaillet1985probabilistic} formally introduced the TSP with stochastic Customers (SS-TSP-C), a.k.a. the probabilistic TSP (PTSP) or TSPSC in the literature, and provided mathematical formulations and a number of properties and bounds of the problem (see also \citeauthor{Jaillet1988}, \citeyear{Jaillet1988}). 
In particular, he showed that an optimal solution for the deterministic problem may be arbitrarily bad in case of uncertainty. 
\cite{laporte1994priori} developed the first exact solution method for the SS-TSP-C, using the integer L-shaped method for two-stage stochastic programs proposed in \cite{LaporteLouveaux1993} to solve instances up to 50 customers.
Heuristics for the SS-TSP-C have then been proposed by 
\cite{jezequel1985probabilistic}, 
\cite{rossi1987aspects}, 
\cite{Bertsimas1988}, 
\cite{bertsimas1993further}, 
\cite{bertsimas1995approaches}, 
\cite{bianchi2005local} and \cite{Bianchi2007} 
as well as meta-heuristics such as simulated annealing (\cite{Bowler2003}) or ant colony optimization (\citeauthor{}{bianchi2002aco}, \citeyear{bianchi2002aco}). 
\cite{Braun2002} proposed a method based on learning theory to approximate SS-TSP-C.
A Pickup and Delivery Traveling Salesman Problem with stochastic Customers is considered in \cite{beraldi2005efficient}, as an extension of the SS-TSP-C in which each pickup and delivery request materializes with a given probability.

Particularly close to the SS-VRPTW-CR is the SS-TSP-C with Deadlines introduced by \cite{Campbell2008x}. 
Unlike the SS-VRPTW-CR, authors assume that customer presences are not revealed at some random moment during the operations, but all at once at the beginning of the day.
However, \citeauthor{Campbell2008x} showed that deadlines are particularly challenging when considered in a stochastic context, and proposed two recourse strategies to address deadline violations. 
\cite{weyland2013improved} later proposed heuristics for the later problem based on general-purpose computing on graphics processing units.
A recent literature review on the SS-TSP-C may be found in \cite{Henchiri2014}. 

The first SS-VRP-C has been studied by \cite{jezequel1985probabilistic}, \cite{jaillet1987stochastic} and \cite{jaillet1988probabilistic} as a generalization of the SS-TSP-C. 
\cite{waters1989vehicle} considered general integer demands and compared different heuristics.
\cite{Bertsimas1992} considered a VRP with stochastic Customers and Demands (SS-VRP-CD). 
A customer demand is assumed to be revealed either when the vehicle leaves the previous customer or when it arrives at the customer's own location. Two different recourse strategies are proposed, as illustrated in Figure \ref{fig:example_bertsimas}. 
For both strategies, closed-form mathematical expressions are provided to compute the expected total distance, provided a first stage solution. 
\cite{gendreau1995exact} and \cite{seguin1994problemes} developed the first exact algorithm for solving the SS-VRP-CD for instances up to 70 customers, by means of an integer L-shaped method. 
\cite{gendreau1996tabu} later proposed a tabu search to efficiently approximate the solution. 
Experimentations are reported on instances with up to 46 customers. 
\cite{gounaris2014adaptive} later developed an adaptive memory programming metaheuristic for the SS-VRP-C and assessed it on benchmarks with up to 483 customers and 38 vehicles.

\cite{Sungur2010} considered a variant of the SS-VRPTW-C, \emph{i.e.}, the Courier Delivery Problem with Uncertainty.
Potential customers have deterministic soft time windows but are present probabilistically, with uncertain service times. 
Vehicles are uncapacitated and share a common hard deadline for returning to the depot. 
The objective is to construct an {\em a priori} solution, to be used every day as a basis for adapting to daily customer requests.
Unlike the SS-VRPTW-CR, the set of customers is revealed at the beginning of the operations.   

\cite{heilporn2011integer} introduced the Dial-a-Ride Problem (DARP) with stochastic customer delays. 
The DARP is a generalization of the VRPTW that distinguishes between pickup and delivery locations and involves customer ride time constraints. 
Each customer is present at its pickup location with a stochastic delay. 
A customer is then skipped if it is absent when the vehicle visits the corresponding location, involving the cost of fulfilling the request by an alternative service (\emph{e.g.}, a taxi). 
In a sense, stochastic delays imply that each request is revealed at some uncertain time during the planning horizon. 
That study is thus related to our problem, although in the SS-VRPTW-CR only a subset of the requests are actually revealed. 
Similarly, \cite{ho2011local} studied a probabilistic DARP where {\em a priori} routes are modified by removing absent customers at the beginning of the day, and proposed local search based heuristics.


\section{Problem description: the SS-VRPTW-CR\label{sec:Problem-description}}

In this section, we recall the description of the SS-VRPTW-CR initially introduced in \cite{Saint-Guillain2017}.

\subsubsection*{Input data.}

We consider a complete directed graph $G=(V,A)$ and a discrete time horizon $H=[1,h]$, where interval $[a,b]$ denotes the set of all integer values $i$ such that $a \le i \le b$. 
To every arc $(i,j) \in A$ is associated a travel time (or distance) $d_{i,j} \in \mathbb{N}$. 
The set of vertices $V=\{0\} \cup W \cup C$ is composed of a depot $0$, a set of $m$ waiting locations $W=[1,m]$ and a set of $n$ customer locations $C=[m+1,m+n]$.
We note $W_0=W \cup \{0\}$ and $C_0=C \cup \{0\}$. 
The fleet is composed of $K$ vehicles of maximum capacity $Q$. 

We consider the set $R=C \times H$ of potential requests such that an element $(c,\Gamma) \in R$ represents a potential request revealed at time $\Gamma \in H$ for customer location $c$.
To each potential request $r=(c, \Gamma) \in R$ is associated a deterministic demand $q_r\in [1,Q]$, a deterministic service duration $s_r \in H$ and deterministic time window $[e_r,l_r]$ with $\Gamma \le e_r \le l_r \le h$. 
We note $p_r$ the probability that $r$ appears on vertex $c$ at time $\Gamma$, and assume independence between request probabilities.

To simplify notations, a request $r=(c,\Gamma)$ may be written in place of its own location $c$. 
For instance, the distance $d_{v,c}$ may also be written $d_{v,r}$.
Furthermore, we use $\Gamma_r$ to denote the reveal time of a request $r \in R$ and $c_r$ for its customer location.
The main notations are summarized in Table \ref{table:notations}.
\begin{table}[h]
\footnotesize
\begin{tabular}{llll}
\hline
$G=(V,A)$ 		& Complete directed graph						&	$r=(c,\Gamma)$ 	& Potential req. $r\in R$ associated	\\
$V=\{0\} \cup W \cup C$	& Set of vertices (depot is $0$)		&								& to time $\Gamma \in H$ and loc. $c \in C$	\\
$W=[1,m]$		& Waiting vertices 								&	$\Gamma_r$					& Reveal time of request $r \in R$	\\
$C=[m+1,m+n]$	& Customer locations 							&	$c_r$						& Cust. loc. hosting req. $r \in R$	\\
$d_{i,j}$ 		& Travel time of arc $(i,j) \in A$				&	$s_r$						& Service time of request $r \in R$	\\	
$K$				& Number of vehicles 		 					&	$[e_r,l_r]$					& Time window of request $r \in R$	\\
$Q$				& Vehicle capacity								&	$q_r$						& Demand of request $r\in R$\\
$H = [1,h]$		& Discrete time horizon		 					&	$p_r$ 						& Prob. associated with req. $r$\\
$R=C \times H$ 	& Set of potential requests	 					&		\\
\hline
\end{tabular}
\caption{\label{table:notations}
Notation summary: graph and potential requests.}
\end{table}

\subsubsection*{First stage solution.}

The first-stage solution is computed offline, before the beginning of the time horizon.
It consists in a set of $K$ vehicle routes visiting a subset of the $m$ waiting vertices, together with time variables denoted $\tau$ indicating how long a vehicle should wait on each vertex. 
More specifically, we denote $(x,\tau)$ a first stage solution to the SS-VRPTW-CR, where:
\begin{itemize}
\item $x = \{ x_1, ..., x_K \}$ defines a set  of $K$ sequences of waiting vertices of $W$. Each sequence $x_k = \langle w_{m_1}, ..., w_{m_k} \rangle$ is such that $x_k$ starts and ends with $0$, {\em i.e.}, $w_{m_1}=w_{m_k}=0$, and each vertex of $W$ occurs at most once in $x$.
We note $W^x \subseteq W$ the set of waiting vertices visited in $x$. 
\item $\tau : W^x \rightarrow H$ associates a waiting time $\tau_w$ to every waiting vertex $w \in W^x$.
\item for each sequence $x_k = \langle w_{m_1}, ..., w_{m_k} \rangle$, the vehicle is back to the depot before the end $h$ of the time horizon, {\em i.e.},
$$\sum_{i=1}^{k-1} d_{w_{m_i},w_{m_{i+1}}} + \sum_{i=2}^{k-1} \tau_{w_{m_i}} \le h $$
\end{itemize}
In other words, $x$ defines a \emph{Team Orienteering Problem} (TOP, see \cite{Chao1996}) to which each visited location is assigned a waiting time by $\tau$.

Given a first stage solution $(x,\tau)$, we define $on(w) = [\underline{on}(w), \overline{on}(w)]$ for each vertex $w \in W^x$ such that $\underline{on}(w)$ (resp. $\overline{on}(w)$) is the arrival (resp. departure) time on $w$.
In a sequence $\langle w_{m_1}, ..., w_{m_k}\rangle$ in $x$, we then have
$\underline{on}(w_{m_i}) = \overline{on}(w_{m_{i-1}}) + d_{w_{m_{i-1}},w_{m_i}}$ and $\overline{on}(w_{m_i}) = \underline{on}(w_{m_i}) + \tau_{w_{m_i}} $
for $i \in [2,k]$ and assume $\overline{on}(w_{m_1}) = 1$.
Figure \ref{fig:partial_second_stage} (left) illustrates an example of first stage solution on a basic SS-VRPTW-CR instance.

\subsubsection*{Recourse strategy and second stage solution.}
A recourse strategy $\mathcal{R}$ states how a second stage solution is gradually constructed as requests are dynamically revealed.
In this paragraph, we define the properties of a recourse strategy. Three recourse strategies are given in Sections \ref{sec:R1_infinite_capacity}, \ref{sec:R_Q} and \ref{sec:R_Q_plus}.

Let $\xi \subseteq R$ be the set of requests that reveal to appear by the end of the horizon $H$. 
The set $\xi$ is also called a \emph{scenario}. 
We note $\xi^t \subseteq \xi$ the set of requests appearing at time $t \in H$, {\em i.e.}, $\xi^t = \{r \in \xi : \Gamma_r = t\}$.
We note $\xi^{1..t} = \xi^1 \cup ... \cup \xi^t$ the set of requests appeared up to time $t$.

A second stage solution is incrementally constructed at each time unit by following the skeleton provided by the first stage solution $(x,\tau)$.
At a given time $t$ of the horizon, we note $(x^t, A^t)$ the current state of the second stage solution:
\begin{itemize}
\item $x^t$ defines a set of vertex sequences describing the route operations performed up to time $t$.
	Unlike $x$, we define $x^t$ on a graph that also includes the customer locations. 
	Sequences of $x^t$ must satisfy the time window and capacity constraints imposed by the VRPTW.
\item  $A^t \subseteq \xi^{1..t}$ is the set of \emph{accepted} requests up to time $t$.
	Requests of $\xi^{1..t}$ that do not belong to $A^t$ are said to be \emph{rejected}.
\end{itemize} 
We distinguish between requests that are accepted and those that are both accepted and satisfied. 
Up to a time $t$, an accepted request is said to be \emph{satisfied} if it is visited in $x^t$ by a vehicle.
Accepted requests that are no yet satisfied must be \emph{guaranteed to be eventually satisfied} according to their time window.

\begin{figure}
\includegraphics[width=.45\textwidth]{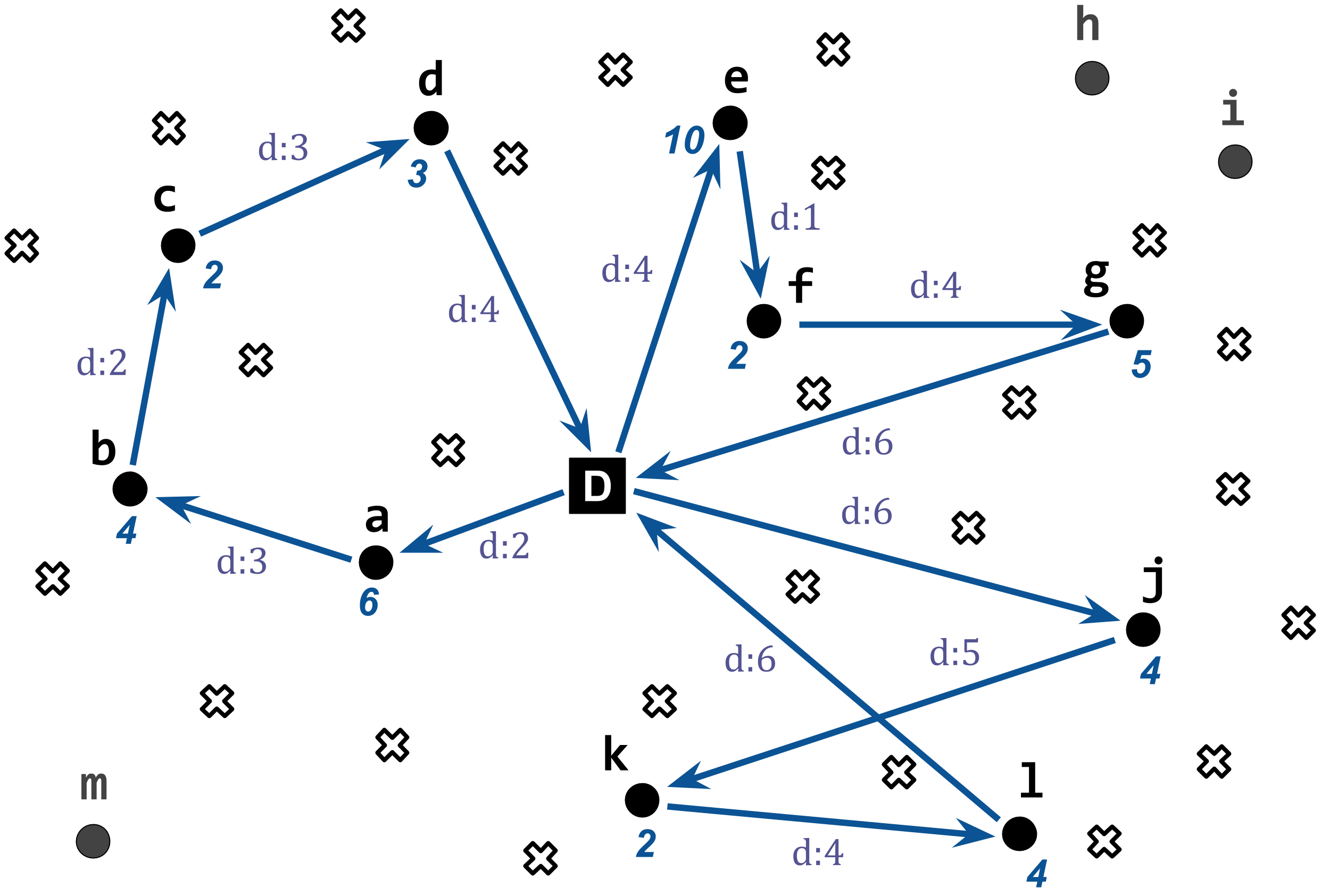}\hfill
\includegraphics[width=.45\textwidth]{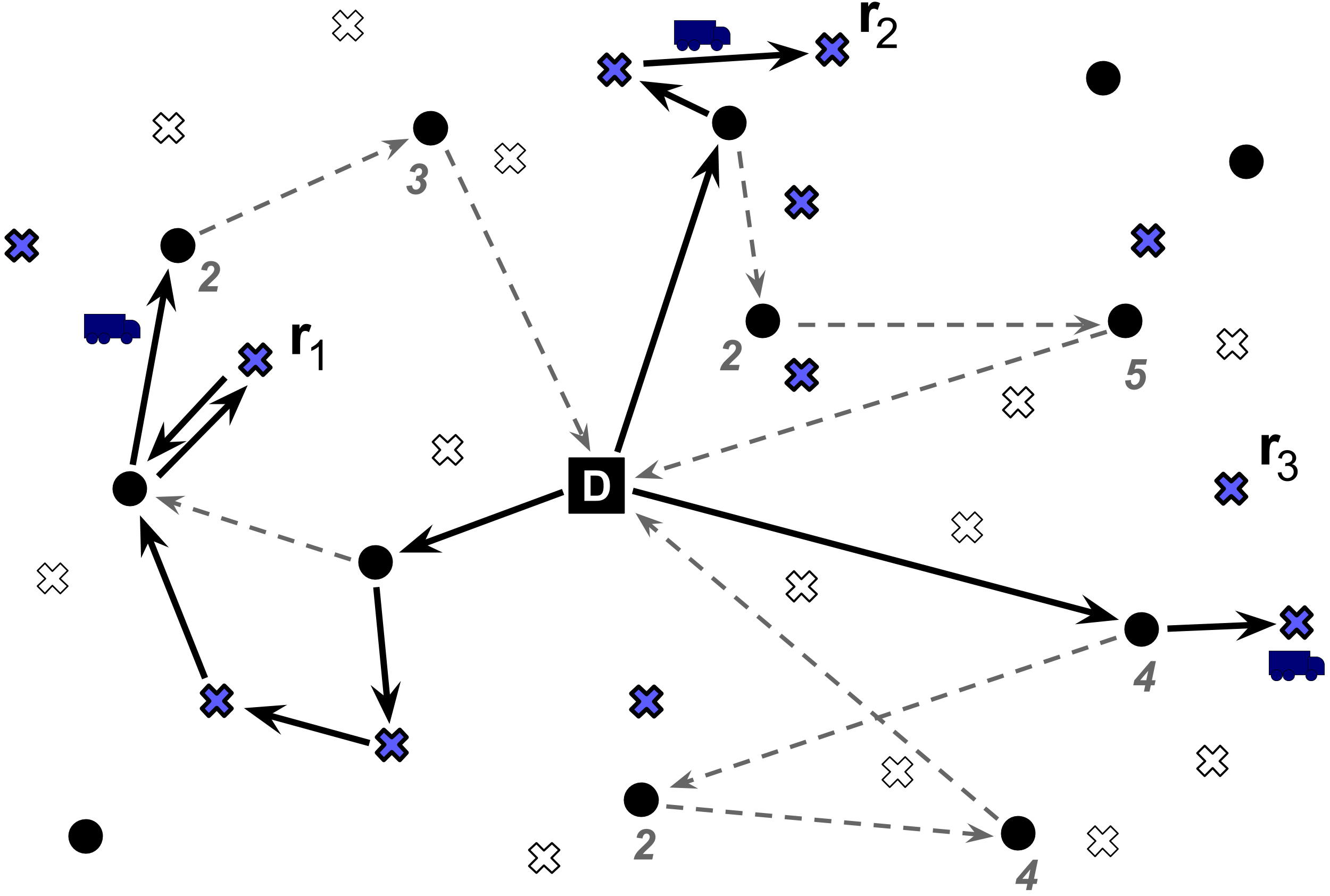}
\caption{\label{fig:partial_second_stage}
On the left:  Example of first stage solution with $K=3$ vehicles.
The depot, waiting vertices and customer locations are represented by a square, circles and crosses, respectively. 
Arrows represent the three vehicle routes: $x_1=\langle D,a,b,c,d,D\rangle$, $x_2=\langle D,e,f,g,D\rangle$ and $x_3=\langle D,j,k,l,D \rangle$. Integers at waiting locations indicate waiting times defined by $\tau$. 
Waiting vertices $h$, $i$ and $m$ are not part of the first stage solution and $W^x=\{a,b,c,d,e,f,g,j,l,k\}$.
For vehicle 1, we have $\overline{on}(D)=1$, $\underline{on}(a)=3$, $\overline{on}(a)=9$, $\underline{on}(b)=12$, $\overline{on}(b)=16$, etc.\newline
On the right: Example of partial second stage solution (plain arrows). 
Filled crosses are accepted requests. 
Some accepted requests, such as $r_1$, have been satisfied (or the vehicle is currently traveling towards the location, {\em e.g.}, $r_2$), while some others are not yet satisfied ({\em e.g.}, $r_3$). 
} 
\end{figure}

Figure \ref{fig:partial_second_stage} (right) illustrates an example of second stage solution being partially constructed at some moment of the time horizon.
\vspace{1em}

\vspace{1em}
Before starting the operations (at time $0$), $x^0$ is a set of $K$ sequences that only contain vertex $0$, 
and $A^0 = \emptyset$. 
At each time unit $t \in H$, given a first stage solution $(x,\tau)$, a previous state $(x^{t-1},A^{t-1})$ of the second stage solution and a set $\xi^{t}$ of requests appearing at time $t$, the new state $(x^t,A^t)$ is obtained by applying a specific recourse strategy $\mathcal{R}$:
\begin{equation}
(x^t,A^t) = \mathcal{R}\big( (x,\tau), (x^{t-1},A^{t-1}), \xi^{t} \big) \label{eq:recourse_computation}  .
\end{equation} 
A necessary property of a recourse strategy is to avoid reoptimization. 
We consider that $\mathcal{R}$ avoids reoptimization if the computation of $(x^t,A^t)$ is achieved in polynomial time. 

We note $\text{cost}(\mathcal{R}, x, \tau, \xi) = | \xi \setminus A^h |$ the final cost of a second stage solution with respect to a scenario $\xi$, given a first stage solution $(x,\tau)$ and under a recourse strategy $\mathcal{R}$. This cost is the number of requests that are rejected at the end $h$ of the time horizon.

\subsubsection*{Optimal first stage solution.}
Given strategy $\mathcal{R}$, an optimal first stage solution $(x,\tau)$ to the SS-VRPTW-CR minimizes the expected cost of the second stage solution:
\begin{align}
\text{(SS-VRPTW-CR)}\hspace{1em} &  \underset{x,\tau}{\text{Minimize}} \hspace{0.5em}  \mathcal{Q^R}(x,\tau) \hspace{1em} \label{ss-vrptw-general} \\
& \text{s.t.}  \hspace{1em} ~(x,\tau) \text{ is a first stage solution}  .  \label{ss-vrptw-general-2} 
\end{align}
The objective function $\mathcal{Q^R}(x,\tau)$, which is nonlinear in general, is the \emph{expected number of rejected requests}, {\em i.e.}, requests that fail to be visited under recourse strategy $\mathcal{R}$ and first stage solution $(x,\tau)$:
\begin{equation}
\mathcal{Q^R}(x,\tau) =  \sum_{\xi \subseteq R} ~ \text{Pr}(\xi) ~ \text{cost}(\mathcal{R}, x, \tau, \xi) \label{eq:expectation_naive}
\end{equation}
where $\text{Pr}(\xi)$ defines the probability of scenario $\xi$. Since we assume independence between requests, we have $\text{Pr}(\xi) = \prod_{r \in \xi} p_r \cdot \prod_{r \in R \setminus \xi} (1 - p_r)$.

\section{Recourse strategy with infinite capacity: $\mathcal{R}^\infty$ \label{sec:R1_infinite_capacity}}

In this section, we recall the recourse strategy introduced in \cite{Saint-Guillain2017} and denoted $\mathcal{R}^\infty$.
It considers the case where vehicles are not constrained by the capacity. In other words, it assumes $Q=\infty$. 
We first describe the recourse strategy $\mathcal{R}^\infty$, and then describe the closed form expression that allows us to compute the expected cost of the second stage solution obtained when applying this recourse strategy to a first stage solution.

\subsection{Description of $\mathcal{R}^\infty$}\label{sec:R_inf_recourse}

In order to avoid reoptimization, each potential request $r \in R$ is assigned to exactly one waiting vertex (and hence, one vehicle) in $W^x$. Informally, the recourse strategy $\mathcal{R}^\infty$ accepts a request revealed at time $t$ if it is possible for the vehicle associated to its corresponding waiting vertex location to adapt its first stage tour to visit the customer, given the set of requests that have been already accepted. The vehicle will then travel from the waiting location to the customer and return to the waiting location.  Time window constraints should be respected, and the already accepted requests should not be perturbed.

\subsubsection*{Request ordering.}
In order to avoid reoptimization, the set $R$ of potential requests must be ordered. This ordering is defined before computing first-stage solutions. Different orders may be considered, without loss of generality, provided that the order is total, strict, and consistent with the reveal time order, {\em i.e.}, $\forall r_1, r_2 \in R$, if the reveal time of $r_1$ is smaller than the reveal time of $r_2$ ($\Gamma_{r_1} < \Gamma_{r_2}$), then $r_1$ must be smaller than $r_2$ in the request order.

In this paper, we order $R$ by increasing reveal time first, end of time window second and lexicographic order to break further ties. More precisely, we consider the order $<_R$ such that $\forall \{r_1,r_2\}\subseteq R$, $r_1<_R r_2$ iff $\Gamma_{r_1} < \Gamma_{r_2}$ or ($\Gamma_{r_1} = \Gamma_{r_2}$ and $l_{r_1}<l_{r_2}$) or ($\Gamma_{r_1} = \Gamma_{r_2}$, $l_{r_1}=l_{r_2}$ and $r_1$ is smaller than $r_2$ according to the lexicographic order defined over $C\times H$).

\subsubsection*{Request assignment according to a first stage solution.}
Given a first-stage solution $(x,\tau)$, we assign each request of $R$ either to a waiting vertex visited in $x$, or to $\bot$ to denote that $r$ is not assigned. We note $\w: R\rightarrow W^x\cup \{\bot\}$ this assignment.
It is computed for each first stage solution $(x, \tau)$ before the application of the recourse strategy. 
To compute this assignment, we first compute for each request $r$ the set $W^x_r$ of waiting locations which are feasible for $r$: 
\[W^x_r = \{ w \in W^x :  t^\text{min}_{r,w} \le t^\text{max}_{r,w}  \} \notag\]
where $t^\text{min}_{r,w}$ and $t^\text{max}_{r,w}$ are defined as follows:
\begin{itemize}
\item $t^\text{min}_{r, w}= \max\{\underline{on}(w), ~ \Gamma_{r}, ~ e_{r} - d_{w,{r}}\}$ is the earliest time for leaving waiting location $w$ to satisfy request $r$. Indeed, a vehicle cannot handle $r$ before (1) the vehicle is on $w$, (2) $r$ is revealed, and (3) the beginning $e_r$ of the time window minus the time $d_{w,{r}}$ needed to go from $w$ to $r$. 
\item $t^\text{max}_{r,w} = \min \{ l_{r} - d_{w,r} , ~ \overline{on}(w) - d_{w,r} - s_{r} - d_{r, w} \}$ is the latest time at which a vehicle can handle $r$ (which also involves a service time $s_r$) from waiting location $w$ and still leave it in time $t \le \overline{on}(w)$.
\end{itemize}
Given the set $W^x_r$ of feasible waiting locations for $r$, we define the waiting location $\w(r)$ associated with $r$ as follows:
\begin{itemize}
\item If $W^x_r = \emptyset$, then $\w(r)=\bot$ ($r$ is always rejected as it has no feasible waiting vertex);
\item If there is only one feasible vertex, {\em i.e.}, $W^x_r = \{ w\}$, then $\w(r)=w$;
\item If there are several feasible vertices, {\em i.e.}, $|W^x_r| > 1$, then $\w(r)$ is set to the feasible vertex of $W^x_r$ that has the least number of requests already assigned to it (break further ties w.r.t. vertex number). This heuristic rule aims at evenly distributing potential requests upon waiting vertices.
\end{itemize}
Once finished, the request assignment ends up with a partition $\{ \pi_\bot, \pi_1, ..., \pi_K \}$ of $R$,
where $\pi_k$ is the set of  requests assigned to the waiting vertices visited by vehicle $k$ and $\pi_\bot$ is the set of unassigned  requests (such that $\w(r)=\bot$).
We note $\pi_w$ the set of  requests assigned to a waiting vertex $w \in W^x$.
We note $\fst(\pi_w)$ and $\fst(\pi_k)$ the first request of $\pi_w$ and $\pi_k$, respectively, according to order $<_R$. For each request $r \in \pi_k$ such that $r \neq \fst(\pi_k)$ we note $\prv(r)$ the request of $\pi_k$ that immediately precedes $r$ according to order $<_R$.

Table \ref{table:notations_2} summarizes the main notations introduced in this section.
Remember that they are all specific to a first stage solution $(x, \tau)$:
\begin{table}[h]
\small
\begin{tabular}{llll}
\hline
$\w(r)$					& Waiting vertex of $W^x$ to which $r \in R$ is assigned (or $\bot$ if $r$ is not assigned)\\
$\pi_k$					& Potential requests assigned to vehicle $k$, \emph{i.e.} ~~~~$\pi_k = \{ r \in R : \w(r) \in x_k \}$		\\
$\pi_w$					& Potential requests assigned to wait. loc. $w \in W^x$, ~~~~$\pi_w = \{ r \in R : \w(r) = w \}$	\\
$\fst(\pi_w)$			& Smallest request of $\pi_w$ according to $<_R$. 		\\
$\fst(\pi_k)$			& Smallest request of $\pi_k$ according to $<_R$. 		\\
$\prv(r)$ 				& Request of $\pi_k$ which immediately precedes $r$ according to $<_R$, if any 	\\
$t^\text{min}_{r,w}$	& Minimum time from which a vehicle can handle request $r \in R$ from $w \in W^x$ \\
$t^\text{max}_{r,w}$	& Maximum time from which a vehicle can handle request $r \in R$ from $w \in W^x$ \\
\hline
\end{tabular}
\caption{\label{table:notations_2}
Notations summary: material for recourse strategies.}
\end{table}

\subsubsection*{Using $\mathcal{R}^\infty$ to adapt a first stage solution at a current time $t$.}

At each time step $t$, the recourse strategy is applied to compute the second stage solution $(x^{t},A^{t})$, given the first stage solution $(x,\tau)$, the second stage solution $(x^{t-1},A^{t-1})$ at the end of time $t-1$, and the incoming requests $\xi^t$.
Recall that $A^{t-1}$ is likely to contain some requests that have been accepted but are not yet satisfied. 

\paragraph{Availability time ${\it available}(r)$}
Besides vehicle operations, a key point of the recourse strategy is to decide, for each request $r\in \xi^t$ that reveals to appear at time $t = \Gamma_r$, whether it is accepted or not. Let $k$ be the vehicle associated with $r$, {\em i.e.}, $r\in\pi^k$.
The decision to accept or not $r$ depends on the time at which $k$ will be available to serve $r$. By available, we mean that it has finished to serve all its accepted requests that precede $r$ (according to $<_R$), and it has reached waiting vertex $\w(r)$. This time is denoted ${\it available}(r)$.
It is defined only when we know all accepted requests that are assigned to $\w(r)$ and that must be served before $r$. As the requests assigned to $\w(r)$ are ordered by increasing reveal time, we know for sure all these accepted requests when $t\geq \Gamma_{\prv(r)}$. In this case, 
${\it available}(r)$ is recursively defined by:
\begin{align}
\small
{\it available}(r) = \begin{cases}
		~~ \underline{on}(\w(r))   & \text{if } ~~ r  = \fst(\pi_{\w(r)})  \\
		~~ {\it available}(\prv(r)) & \text{if } ~~ r  \neq \fst(\pi_{\w(r)})\\
        &\mbox{and }\prv(r) \notin A^t  \\
		~~ \max \{e_{\prv(r)}, {\it available}(\prv(r)) + d_{\w(r),\prv(r)} \}&\\ 
        ~~~~~+ ~ s_{\prv(r)} + d_{\prv(r), \w(r)}  & \text{otherwise} . 	
	\end{cases} \notag
\end{align}

\paragraph{Request notifications}

$A^t$ is the set of appeared requests accepted up to time $t$. It is initialized with $A^{t-1}$ as all requests previously accepted must still be accepted at time $t$. Then, each incoming request $r\in \xi^t$ is considered, taken by increasing order with respect to $<_R$. $r$ is either accepted (added to $A^t$) or rejected (not added to $A^t$) by applying the following decision rule:
\begin{itemize}
\item If $\w(r)=\bot$ or ${\it available}(r) > t^\text{max}_{r,\w(r)}$ then $r$ is rejected;
\item Else $r$ is accepted and added to $A^t$; 
\end{itemize}

\paragraph{Vehicle operations}

\begin{figure}
\centering
\includegraphics[width=1.0\textwidth]{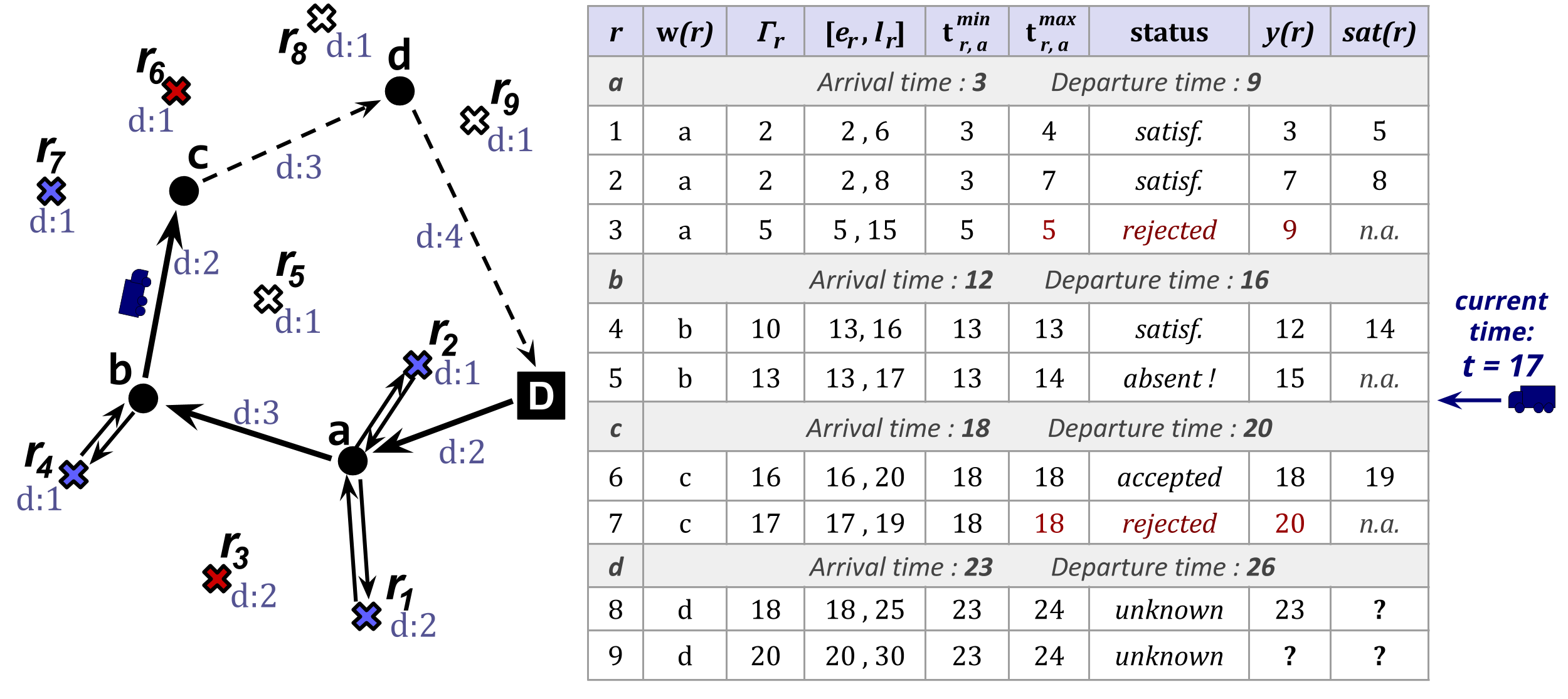}
\caption{
Example of second stage solution at time $t=17$, under strategy $\mathcal{R}^\infty$. 
A filled cross represents a request that appeared, an empty one a request that is either still unknown ({\em e.g.}, $r_8$) or revealed as being absent ({\em i.e.}, did not appear, {\em e.g.}, $r_5$). 
Here $\pi_k=\langle r_a, r_1, \dots, r_9 \rangle$ is the sequence of requests assigned to the vehicle, according to $(x,\tau)$.
We assume $q_r = s_r = 0, \forall r \in R$.
$sat(r)$ represents, for a request $r$, the time at which $r$ gets satisfied.
}
\label{fig:instance_R1}
\end{figure}

Once $A^t$ has been computed, vehicle operations for time unit $t$ must be decided. 
Vehicles operate independently to each other.
If vehicle $k$ is traveling between a waiting location and a customer location, or if it is serving a request, then its operation remains unchanged;
Otherwise, let $w$ be the current waiting location (or the depot) of vehicle $k$:
\begin{itemize}
\item 	If $t = \overline{on}(w)$, the operation for $k$ is "travel from $w$ to the next waiting vertex (or the depot), as defined in the first stage solution";
\item 	Otherwise, let $P= \{ r \in \pi_w  |  c_r \notin x^t  \wedge  (r \in A^t \vee t < \Gamma_r )\}$ be the set of requests of $\pi_w$ that are not yet satisfied and that are either accepted or with unknown revelation.
\begin{itemize}
\item If $P=\emptyset$, then the operation for $k$ is "travel from $w$ to the next waiting vertex (or the depot), as defined in the first stage solution";
\item Otherwise, let $r^\text{next}$ be the smallest element of $P$ according to $<_R$
\begin{itemize}
\item If $t < t^\text{min}_{r^\text{next},w}$, then the operation for $k$ is "wait until $t+1$";
\item Otherwise, the operation is "travel to $r^\text{next}$, serve it and come back to $w$".
\end{itemize}
\end{itemize}
\end{itemize}
Figure \ref{fig:instance_R1} shows an example of second stage solution at a current time $t=17$, from an operational point of view. \\

\subsection{Expected cost of second stage solutions under $\mathcal{R}^\infty$ \label{sec:expected_cost_R_infinity}}

Provided a recourse strategy $\mathcal{R}$ and a first stage solution $(x,\tau)$ to the SS-VRPTW-CR, 
a naive approach for computing  $\mathcal{Q^R}(x,\tau)$ would be to literally follow equation (\ref{eq:expectation_naive}), therefore using the strategy described by $\mathcal{R}$ in order to confront $(x,\tau)$ to each and every possible scenario $\xi \subseteq R$.
Because there is an exponential number of scenarios with respect to $|R|$, this naive approach is not affordable in practice.
In this section, we show how the expected number of rejected requests under the recourse strategy $\mathcal{R}^\infty$ may be computed in $\mathcal{O}(nh^2)$ using closed form expressions.

Let us recall that we assume that the potential request probabilities are independent from each other such that, for any couple of requests $r, r' \in R$, the probability $p_{r \wedge r'}$ that both requests appear is given by $p_{r \wedge r'} = p_r \cdot p_{r'}$.

$\mathcal{Q^{R^\infty}}(x,\tau)$ is equal to the expected number of rejected requests, which in turn is equal to the expected number of requests that reveal to appear minus the expected number of accepted requests. The expected number of revealed requests is given by the sum of all request probabilities, whereas the expected number of accepted requests is equal to the sum, for every request $r$, of the probability that it belongs to $A^h$, {\em i.e.}, 
\begin{equation}
\mathcal{Q}^{\mathcal{R}^\infty}(x,\tau) =  \sum_{r \in R} p_r - \sum_{r \in R} \text{Pr}\{ r \in A^h \} 
								= \sum_{r \in R} \big( p_r - \text{Pr}\{ r \in A^h \} \big)	\label{eq:expectation_infinite}
\end{equation}
where the right-hand side of the equation comes from the independence hypothesis. 

Given a request $r \in \pi_k$, the probability $\text{Pr}\{ r \in A^h \}$ depends on (1) whether $r$ reveals to appear or not, and (2) the time at which vehicle $k$ leaves $\w(r)$ to satisfy $r$ if condition (1) is satisfied. It may be computed by considering all time units for which vehicle $k$ can leave $\w(r)$ to satisfy request $r$:
\begin{align}
\text{Pr}\{ r \in A^h \} & = \sum_{t=t^\text{min}_{r, \w(r)}}^{t^\text{max}_{r, \w(r)}} g_1(r, t)  
\label{eq:req-satisfiable_infinite} 
\end{align}

\paragraph{Computation of probability $g_1(r,t)$}
$g_1(r,t)$ is the probability that $r$ has been revealed and vehicle $k$ leaves $\w(r)$ at time $t$ to serve $r$.
Let us note $\departureTime(r) = \max\{{\it available}(r), t^\text{min}_{r,\w(r)}\}$ the time at which vehicle $k$ actually leaves the waiting vertex $\w(r)$ in order to serve $r$ (the vehicle may have to wait if ${\it available}(r)$ is smaller than the earliest time for leaving $\w(r)$ to serve $r$).
The probability $g_1(r,t)$ is defined by:
\begin{align}
g_1(r,t) \equiv \text{Pr}\{ & r\in\xi^{\Gamma_r}  \text{ and } \departureTime(r) = t \} \notag .
\end{align}
This probability is computed only when $t^\text{min}_{r, \w(r)}\leq t\leq t^\text{max}_{r, \w(r)}$.
In this case, the reveal time $\Gamma_r$ of $r$ is passed ({\em i.e.}, $t\geq \Gamma_r$). 

Since $\departureTime(r)$ depends on previous operations on waiting location $\w(r)$, we compute $g_1$ recursively starting from the first request $\fst(\pi_{\w(r)})$ assigned to $\w(r)$, up to the last request assigned to $\w(r)$.
The second stage solution strictly respects the first stage schedule when visiting the waiting vertices, that is, vehicle $k$ first reaches $\w(r)$ at time $\underline{on}(\w(r))$ and last leaves it at time $\overline{on}(\w(r))$. 

The base case to compute $g_1$ is: 
\begin{align}
\text{if }r = \fst(\pi_{\w(r)})\text{ then } g_1(r,t) = 
	\begin{cases}
		p_{r} & \text{if } ~~ t = \max\{ \underline{on}(\w(r)),  t^\text{min}_{r, \w(r)} \} \\
		0  & \text{otherwise}.
	\end{cases} \label{eq:g1_basic_base_infinite}
\end{align}
Indeed, if the first request $r$ assigned to a waiting vertex $w$ appears then it is necessarily accepted and vehicle $k$ leaves $w$ to serve $r$ either at time $\underline{on}(w)$, or at time $t^\text{min}_{r,w}$ if 
the time window for serving $r$ begins after $\underline{on}(w) + d_{w,r}$.

The general case of a request $r$ which is not the first request assigned to $\w(r)$ ({\em i.e.}, $r \neq \fst(\pi_{\w(r)})$) depends on the time ${\it available}(r)$ at which vehicle $k$ becomes available for $r$.
Whereas ${\it available}(r)$ is deterministic when we know the set $A^{\Gamma_{\prv(r)}}$ of previously accepted requests, it is not anymore in the context of the computation of probability $g_1(r,t)$. We are thus interested in its probability distribution $f(r,t)\equiv Pr \{ {\it available}(r) = t\}$. The computation of $f$ is detailed below.
Given $f$, the general case to compute $g_1$ is:
\begin{align}
\text{if }r \neq \fst(\pi_{\w(r)})\text{ then }g_1(r,t) = 
	\begin{cases}
		p_{r} \cdot f(r,t)   & \text{if } t > t^\text{min}_{r, \w(r)}  \\
		p_{r} \cdot \sum_{t' = \underline{on}(w)}^{t^\text{min}_{r, \w(r)}} 
			f(r,t') & \text{if } t = t^\text{min}_{r,\w(r)}\\
		0 & \text{otherwise} . 			
	\end{cases}  \label{eq:g1_basic_infinite}
\end{align}
Indeed, if $t > t^\text{min}_{r,\w(r)}$, then vehicle $k$ leaves $\w(r)$ to serve $r$ as soon as it becomes available. 
If $t < t^\text{min}_{r,\w(r)}$, the probability that vehicle $k$ leaves $\w(r)$ at time $t$ is null since $t^\text{min}_{r,\w(r)}$ is the earliest time for serving $r$ from location $\w(r)$. 
Finally, at time $t = t^\text{min}_{r,\w(r)}$, we must consider the possibility that vehicle $k$ was waiting for serving $r$ since an earlier time $t' \leq  t^\text{min}_{r,\w(r)}$. 
In this case, the probability that vehicle $k$ leaves $\w(r)$ for serving $r$ at time $t$ is $p_{r}$ times the probability that vehicle $k$ was actually available from a time $\underline{on}(w) \le t' \le t^\text{min}_{r,w}$.

\paragraph{Computation of probability $f(r,t)$}
Let us now define how to compute $f(r,t)$ which is the probability that vehicle $k$ becomes available for $r$ at time $t$, {\em i.e.},
\[f(r,t)\equiv Pr \{ {\it available}(r) = t\}\]
$f(r,t)$ is computed only when $r$ is not the first request assigned to $\w(r)$, according to the order $<_R$, and its computation depends on what happened to the previous request $r_p = \prv(r)$. We have to consider three cases: (1) $r_p$ appeared and got satisfied, (2) $r_p$ appeared but wasn't satisfiable, and (3) $r_p$ didn't appear.

\begin{figure}[t]
\centering
\includegraphics[width=0.8\textwidth]{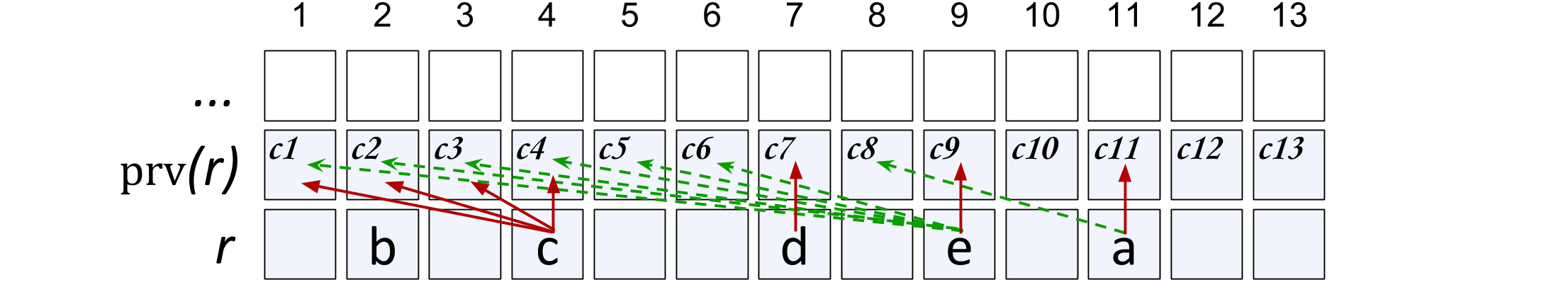}
\caption{
Cell dependencies in a $f(r,t)$-matrix, assuming $\underline{on}(w)=1$, $\Gamma_r = 4$, $t^\text{min}_{r,w} = 6$ and $S_r = 3$. 
Cell \textbf{a} represents $f(r,11)$, whereas cell \textit{\textbf{c8}} represents $f(\prv(r),9)$.
Since $t \ge \Gamma_r$ and $t \ge t^\text{min}_{r,w}$ at cell \textbf{a}, $f(r,11)$ depends directly and exclusively on cells \textit{\textbf{c8}} for its first term of eq. (\ref{eq:h_infinite}), and \textit{\textbf{c11}} for the second and third ones. 
Since cell \textbf{b} has $t < \Gamma_r$ and $t < t^\text{min}_{r,w}$, it does not depend on another cell as its probability must be zero.
Cell \textbf{c} has $t= \Gamma_r$ and $t < t^\text{min}_{r,w}$, so the first term of eq. (\ref{eq:h_infinite}) is zero and the second and third depend on probabilities at cells \textit{\textbf{c1}} to \textit{\textbf{c4}}.
Cell \textbf{d} has $\Gamma_r \le t < t^\text{min}_{r,w}$, so it only depends on cell \textit{\textbf{c7}}.
Finally, since cell \textbf{e} has $t = t^\text{min}_{r,w}$ first term of eq. (\ref{eq:h_infinite}) depends on cells \textit{\textbf{c1}} to \textit{\textbf{c6}}, whereas $t \ge \Gamma_r$ makes second and third terms depend on cell \textit{\textbf{c9}} only.
}
\label{fig:h_matrix}
\end{figure}

For conciseness let us denote by $S_{r_p} = d_{\w(r_p),r_p} + s_{r_p} + d_{r_p, \w(r_p)}$ the amount of time needed to travel from $\w(r_p)$ to $r_p$, serve $r_p$ and travel back to $\w(r_p)$. 
Let also $\delta^w(r, t)$ return $1$ iff request $r$ is satisfiable from time $t$ and vertex $w$, {\em i.e.}, $\delta^w(r, t) = 1$ if $t  \le t^\text{max}_{r,w}$, and $\delta^w(r, t) = 0$ otherwise.
Then:
\begin{align}
f(r,t) = ~ 	g_1(r_p,  t  - S_{r_p}) \cdot \delta^w(r_p, t  - S_{r_p}) + ~ g_1(r_p,  t ) \cdot \big( 1-\delta^w(r_p, t) \big) ~ + ~  g_2(r_p,t)    \label{eq:h_infinite}
\end{align}
where $g_2(r,t)$ is the probability that $r$ did not appear and is discarded at time $t$ (the computation of $g_2$ is detailed below).
The first term in the summation of the right hand side of equation (\ref{eq:h_infinite}) gives the probability that request $r_p$ actually appeared and got satisfied (case 1). 
In such a case, $\departureTime(r_p)$ must be the current time $t$, minus delay $S_{r_p}$ needed for serving $r_p$.
The second and third terms of equation (\ref{eq:h_infinite}) add the probability that the vehicle was available time $t$, but that request $r_p$ did not consume any operational time.
There are only two possible reasons for that: either $r_p$ actually appeared but was not satisfiable (case 2, corresponding to the second term), or $r_p$ did not appear at all (case 3, corresponding to the third term).
Figure \ref{fig:h_matrix} shows how the $f$-probabilities of a request $r$ depend on those of $\prv(r)$. 

\paragraph{Computation of probability $g_2(r,t)$}
Let us finally define how to compute $g_2(r,t)$, which is the probability that a request $r$ did not appear ($r \notin \xi^{\Gamma_r}$) and is discarded at time $t$. Let us note $\text{discardedTime}(r) = \max\{{\it available}(r),\Gamma_r\}$ the time at which the vehicle becomes available for $r$ whereas $r$ does not appear.
\begin{align}
g_2(r,t) \equiv Pr\{r\not\in\xi^{\Gamma_r}\text{ and }t = \text{discardedTime}(r)\}
\end{align}
This probability is computed recursively, like for $g_1$.
The base case is:
\begin{align}
\text{if }r=\fst(\pi_{\w(r)})\text{ then }g_2(r,t) &= 
	\begin{cases}
		1 - p_{r}    & \text{if } ~~ t = \max(\underline{on}(\w(r)), \Gamma_{r} ) \\
		0 & \text{otherwise}
	\end{cases} \label{eq:g2_basic_base_infinite}
\end{align}
The general case is quite similar to the one of function $g_1$. 
We just consider the probability $1- p_{r}$ that $r$ does not reveal and replace $t^\text{min}_{r,\w(r)}$ by $\Gamma_{r}$:\\
\begin{align}
&\mbox{if } r \neq \fst(\pi_{\w(r)}) \mbox{ then }\notag\\ 
&g_2(r,t) = 
	\begin{cases}
		(1 - p_{r}) \cdot f(r,t )   & \text{if } ~~ t > \max(\underline{on}(\w(r)),\Gamma_{r})  \\
		(1 - p_{r}) \cdot \sum_{t'= \underline{on}(\w(r))}^{\max(\underline{on}(\w(r)),\Gamma_{r})} 
			f(r,t' ) ~~ & \text{if } ~~ t = \max(\underline{on}(\w(r)),\Gamma_{r}) \\
		0 & \text{otherwise.}
	\end{cases}  \label{eq:g2_basic_infinite}
\end{align}

\subsubsection*{A note on implementation.}
Since we are interested in computing $\text{Pr}\{ r \in A^h \}$ for each request $r$ separately, by following the definition of $g_1$, we only require the $f$-probability associated to $r$ to be already computed. 
This suggests a dynamic programming approach. 
Computing all the $f$-probabilities can then be incrementally achieved while filling up a 2-dimensional matrix containing all the $f$-probabilities.
By using an adequate data structure while filling up such a sparse matrix, substantial savings can be made on the computational effort. 

\subsubsection*{Computational complexity.}
The time complexity of computing the expected cost is equivalent to the one of filling up a $|\pi_w| \times h$ matrix for each visited waiting location $w \in W^x$, in order to store all the $f(r,t)$ probabilities. 
By processing incrementally on each waiting location separately, each matrix cell can be computed in constant time using equation (\ref{eq:h_infinite}). 
In particular, once the probabilities in cells $(\prv(r),1\cdots t)$ are known, the cell $(r,t)$ such that $r \ne \fst(\pi_w)$ may be computed in $\mathcal{O}(1)$ according to equations (\ref{eq:g1_basic_infinite}) - (\ref{eq:g2_basic_infinite}). 
Given $n$ customer locations and a time horizon of length $h$, we have at most $|R| = nh \ge \sum_{w \in W^x} |\pi_w|$ potential requests.  
It then requires at most $\mathcal{O}(|R|h) = \mathcal{O}(nh^2)$ constant time operations to compute $\mathcal{Q}^{\mathcal{R}^\infty}(x,\tau)$.

\section{Recourse strategy with bounded capacity: $\mathcal{R}^q$ \label{sec:R_Q}}

In this section we show how to generalize $\mathcal{R}^\infty$ in order to handle capacitated vehicles in the context of integer customer demands. 

\subsection{Description of $\mathcal{R}^q$}

The recourse strategy $\mathcal{R}^q$ only differs from $\mathcal{R}^\infty$ in the way requests are either accepted or rejected.
The \emph{request notification} phase (described in Section \ref{sec:R_inf_recourse}) is modified by adding a new condition that takes care of the current vehicle load. Under $\mathcal{R}^q$, a request $r$ is therefore accepted if and only if:
\begin{equation}
\w(r)\neq\bot ~~ \wedge ~~ {\it available}(r) \le t^\text{max}_{r,\w(r)} ~~ \wedge ~~  q_r + \sum_{r' \in \pi_k \cap A^t} q_{r'}  \le Q
\label{eq:conditions_R_capa}
\end{equation}
In Section \ref{sec:R_Q_expectation}, we show how to adapt the closed form expressions described in Section \ref{sec:expected_cost_R_infinity} in order to compute the exact expected cost when vehicles have limited capacity. 
In section \ref{sec:R_Q_ss-vrp-c}, we show how the resulting equations, when specialized to the special case of the SS-VRP-C, naturally reduce to the ones proposed in \cite{Bertsimas1992}.

\subsection{Expected cost of second stage solutions under $\mathcal{R}^q$ \label{sec:R_Q_expectation}}

Contrary to strategy $\mathcal{R}^\infty$, under strategy $\mathcal{R}^q$ a request may be rejected if the corresponding vehicle is full. 
%
Recall that $\pi_k$ is the set of potential requests in route $k \in [1,K]$, ordered by $<_R$.
We note $\load(k,t)$ the load of vehicle $k\in [1,K]$ at time $t\in H$.
The probability $\text{Pr}\{ r \in A^h \}$ now depends on what happened earlier, not only on the current waiting location, but also on previous waiting locations (if any). 
To describe $\text{Pr}\{ r \in A^h \}$, we need to consider every possible \emph{time and load configurations} for $r$:
\begin{align}
\text{Pr}\{ r \in A^h \} &= \sum_{t = t^\text{min}_{r, w}}^{t^\text{max}_{r, w}} \sum_{q=0}^{Q-q_r} g_1(r, t, q) \label{eq:req-satisfiable_capacity} .
\end{align}

\paragraph{Computation of probability $g_1(r,t,q)$}
$g_1(r,t,q)$ is the probability that $r$ has been revealed, and vehicle $k$ leaves $\w(r)$ at time $t$ with load $q$ to serve $r$, {\em i.e.}, 
\begin{align}
g_1(r,t, q) \equiv \text{Pr}\{ &r \in \xi^{\Gamma_r}, \departureTime(r)=t \text{ and }\load(k,t)=q\} \notag 
\end{align}
The base case for computing $g_1$ is now concerned with the very first potential request on the entire route, $r=\fst(\pi_k)$, which must be considered as soon as the vehicle arrives at $w = \w(r)$, that is at time $\underline{on}(w)$ except if 
$\underline{on}(w) < t^\text{min}_{r, w}$:
\begin{align}
\mbox{if } &r = \fst(\pi_k) \mbox{ then }\notag\\
&g_1(r,t,q) = 
	\begin{cases}
		p_{r} & \text{if } ~~ t = \max\{ \underline{on}(w),  t^\text{min}_{r, w} \} ~~\wedge~~ q = 0 \\ 
		0  & \text{otherwise}.
	\end{cases} \label{eq:g1_basic_base}
\end{align}
For any $q \ge 1$, function $g_1(r,t,q)$ must be equal to zero as the vehicle necessarily carries an empty load when considering $r$.

Like under strategy $\mathcal{R}^\infty$, the first potential request $r = \fst(\pi_w)$ of a planned waiting location $w \in W^x$ which is not the first of the route ({\em i.e.}, $r \ne \fst(\pi_k)$) is special as well. 
As the arrival time on $w$ is fixed by the first stage solution, $\departureTime(r)$ is necessarily $\max( \underline{on}(w),  t^\text{min}_{r, w} )$. 
Unlike $\departureTime(r)$, $load(k,t)$ is not deterministic but rather depends on what happened on the previous waiting location $\w(\prv(r))$. Hence, we extend the definition of probability $f(r,t,q)$ to handle this special case: When $r=\fst(\w(r))$, $f(r,t,q)$ is the probability that vehicle $k$ finishes to serve $\prv(r)$ at time $t$ with load $q$. The complete definition and computation of $f$ is detailed in the next paragraph.
Given this probability $f$, the probability that vehicle $k$ carries a load $q$ is obtained by summing these $f$ probabilities for every time unit $t'$ during which vehicle $k$ may serve $\prv(r)$, {\em i.e.} $\underline{on}(\w(\prv(r)))\leq t'\leq \overline{on}(\w(\prv(r)))$.
Therefore, for every request which is the first of a waiting vertex, but not the first of the route, we have:
\begin{align}
\mbox{if } &r=\fst(\pi_{\w(r)}) \mbox{ and }r \neq \fst(\pi_k) \mbox{ then }\notag\\ 
&g_1(r,t,q) = \begin{cases}
		p_{r} \cdot \sum_{t' = \underline{on}(\w(\prv(r))}^{ \overline{on}(\w(\prv(r)))} f(r,t',q) , & \text{if } ~~ t = \max( \underline{on}(w),  t^\text{min}_{r, w} )  \\ 
		0  & \text{otherwise}.
	\end{cases}, \label{eq:g1_basic_base_2}
\end{align}

Finally, the general case of a request $r$ which is not the first request of a waiting location depends on the time and load configuration at which the vehicle is available for $r$:
\begin{align}
\mbox{if } &r \neq \fst(\pi_{\w(r)}) \mbox{ then }\notag\\ 
&g_1(r,t, q) = \begin{cases}
		p_{r} \cdot f(r,t, q)   & \text{if } ~~ t > t^\text{min}_{r,w}  \\
		p_{r} \cdot \sum_{t' = \underline{on}(w)}^{t^\text{min}_{r,w}} 
			f(r,t' , q) ~~~ & \text{if } ~~ t = t^\text{min}_{r,w}  \\
		0 & \text{otherwise} . 	
	\end{cases} \label{eq:g1_R_capa}
\end{align}
If $t > t^\text{min}_{r,w}$, the vehicle leaves $w$ to serve $r$ as soon as it becomes available. 
At $t = t^\text{min}_{r,w}$, we must take into account the possibility that the vehicle was actually available from an earlier time $\underline{on}(w) \le t' < t^\text{min}_{r,w}$.

\paragraph{Computation of probability $f(r,t,q)$}
The definition of the probability $f(r,t,q)$ depends on whether $r$ is the first request assigned to a waiting vertex or not. If $r\neq \fst(\w(r))$, then $f(r,t,q)$ is the probability that vehicle $k$ is available for $r$ at time $t$ with load $q$:
\[f(r,t,q) \equiv \text{Pr}\{ {\it available}(r)=t \text{ and }\load(k,t)=q\}\]
whereas if $r=\fst(\w(r))$, then $f(r,t,q)$ is the probability that vehicle $k$ finishes to serve $\prv(r)$ at time $t$ with load $q$:
\[f(r,t,q) \equiv \text{Pr}\{ {\it finishToServe}(\prv(r))=t \text{ and }\load(k,t)=q\}\]

In both cases, the computation of $f(r,t,q)$ depends on what happened to previous request $r_p=\prv(r)$, and we extend eq. (\ref{eq:h_infinite}) in a straightforward way: We add a third parameter $q$ and remove $q_{r_p}$ load units in the first term, corresponding to the case where $r_p$ appeared and got satisfied:
\begin{align}
f(r,t,q) = ~ 	& g_1(r_p,  t  - S_{r_p}, q - q_{r_p}) \cdot \delta^w(r_p, t  - S_{r_p}, q-q_{r_p}) \notag \\
	    	& + ~ g_1({r_p},  t , q) \cdot \big( 1-\delta^w({r_p}, t,q) \big) ~ + ~  g_2({r_p},t,q)  .  \label{eq:f_capa} 
\end{align}
where $\delta^w(r, t, q)$ returns $1$ iff request $r$ is satisfiable from time $t$, load $q$, and vertex $w$, {\em i.e.}, $\delta^w(r, t, q) = 1$ if $t  \le t^\text{max}_{r,w}$ and $q+q_r\leq Q$, whereas $\delta^w(r, t, q) = 0$ otherwise.

\paragraph{Computation of probability $g_2(r,t,q)$} 
The definition of $g_2$ is
\begin{align}
g_2(r,t,q) &\equiv \text{Pr} \{ r\not\in\xi^{\Gamma_r}, \discardedTime(r)=t \text{ and }\load(k,t)=q\} \notag
\end{align}
The computation of $g_2$ is adapted in a rather straightforward way from eq. (\ref{eq:g2_basic_base_infinite}) and (\ref{eq:g2_basic_infinite}), simply adding the load dimension $q$. 
For the very first request of the route of vehicle $k$, we have:
\begin{align}
\mbox{if } &r = \fst(\pi_k) \mbox{ then }\notag\\ 
&g_2(r,t,q) = \begin{cases}
		1 - p_{r}, & \text{if } ~~ t = \max( \underline{on}(\w(r)), \Gamma_{r}) ~~\wedge~~ q = 0 \\ 
		0  & \text{otherwise}.
	\end{cases} \label{eq:g2_basic_base}
\end{align}
For the first request of a waiting location which is not the first of the route, we have:
\begin{align}
\mbox{if } &r = \fst(\pi_{\w(r)})\mbox{ and }r\neq \fst(\pi_k) \mbox{ then }\notag\\ 
&g_2(r,t,q) = \begin{cases}
		(1 - p_{r}) \cdot \sum_{t' = \underline{on}(\w(\prv(r)))}^{ \overline{on}(\w(\prv(r)))} f(\prv(r),t',q) , & \text{if } ~~ t = \max( \underline{on}(w),  \Gamma_{r} )  \\ 
		0  & \text{otherwise}.
	\end{cases} \label{eq:g2_basic_base_2}
\end{align}
For a request which is not the first of a waiting vertex: 
\begin{align}
\mbox{if } &r \neq \fst(\pi_{\w(r)}) \mbox{ then }\notag\\ 
&g_2(r,t, q) = \begin{cases}
		(1 - p_{r}) \cdot f(\prv(r),t, q)   & \text{if } ~~ t > \max( \underline{on}(w),  \Gamma_r )   \\
		(1 - p_{r}) \cdot \sum_{t' = \underline{on}(w)}^{\max( \underline{on}(w),  \Gamma_r ) } 
			f(\prv(r),t' , q) ~~~ & \text{if } ~~ t = \max( \underline{on}(w),  \Gamma_r )  \\
		0 & \text{otherwise} . 	
	\end{cases} \label{eq:g2_R_capa}
\end{align}

\subsubsection*{Computational complexity}
In the case of capacitated vehicles, the complexity of computing $\mathcal{Q}^{\mathcal{R}^q}(x,\tau)$ is equivalent to the one of filling up $K$ matrices of size $|\pi_k| \times h \times Q$, containing all the $f(r,t,q)$ probabilities. 
Like in strategy $\mathcal{R}^\infty$, by processing incrementally, each cell of each tri-dimensional matrix is computable in constant time.
Given $n$ customer locations and a time horizon of length $h$, there are at most $|R| = nh \ge \sum_{k=1}^K |\pi_k|$ potential requests in total, leading to an overall complexity of $\mathcal{O}(nh^2Q)$.

\subsection{Relation with SS-VRP-C  \label{sec:R_Q_ss-vrp-c}}
As presented in section \ref{sec:Introduction}, the SS-VRP-C differs by having stochastic binary demands, representing the random customer presence, and no time window. 
The objective here minimizes the expected distance traveled, provided that when a vehicle reaches its maximal capacity, it unloads by making a round trip to the depot.
In order to compute the expected length of a first stage solution that visits all the customers, a key point is to compute the probability distribution of the vehicle's current load when reaching a customer. 
In fact, the later is directly related to the probability that the vehicle makes a round trip to the depot to unload, which is denoted by the function ``$f(m,r)$'' in \cite{Bertsimas1992}.
Here we highlight the relation between the SS-VRPTW-CR and SS-VRPTW-C by showing how our equations, when time windows are not taken into account, can be derived to obtain the ``$f(m,r)$'' one proposed in \cite{Bertsimas1992}.

Since there is no time window consideration, we can state $\Gamma_r = t^\text{min}_{r,w} = 1$ and $t^\text{max}_{r,w} = +\infty$ for any request $r$. Also, each demand $q_r$ is equal to 1.
Consequently, the $\delta$-function used in the computation of the $f$ probabilities depends only on $q$ and is equal to 1 if $q \le Q$. Therefore, the $f$ probabilities are defined by:
\begin{align}
f(r,t,q)  = g_1(r_p, t  - S_{r_p}, q-1) ~ + ~ g_2(r_p,  t ,q)   \notag
\end{align}
with $r_p=\prv(r)$.
Now let $f'(r,q)  = \sum_{t \in H} f(r,t,q)$.
As $f(r,t,q)$ is the probability that the vehicle is available for $r$ at time $t$ with load $q$, $f'(r,q)$ is the probability that the vehicle is available for $r$ with load $q$ during the day.
It is also true that $f'(r,q)$ gives the probability that exactly $q$ requests among the $r_1, ..., r_p$ potential ones actually appear (with a unit demand).
We have:
\begin{align}
f'(r,q)  = \sum_{t \in H} f(r,t,q) = \sum_{t \in H} g_1(r_p, t  - S_{r_p}, q-1) ~ + ~ \sum_{t \in H} g_2(r_p,  t ,q) \notag 
\end{align}
As we are interested in $f'(r,q)$, not in the travel distance, let us assume that all the potential requests are assigned to the same waiting location.
Then either $r = \fst(\pi_k)$ or $r \ne \fst(\pi_{\w(r)})$. 
If $r = \fst(\pi_k)$ we naturally obtain:
\begin{align}
f'(r,q)  	&= \sum_{t \in H} p_{r} \cdot f(r, t , q-1) +  \sum_{t \in H} (1 - p_{r}) \cdot f(r, t, q) \notag \\
		&= \begin{cases}
		p_r + (1 - p_r) = 1, & \text{if } ~~ q = 0 \\ 
		0,  & \text{otherwise}.
	\end{cases} \notag
\end{align}
If $r \ne \fst(\pi_{\w(r)})$, since we always have $t \ge t^\text{min}_{r,w}$, we have:
\begin{align}
f'(r,q)  	&= \sum_{t \in H} p_{r} \cdot f(r, t , q-1) +  \sum_{t \in H} (1 - p_{r}) \cdot f(r, t, q) \notag \\
		&= p_{r} \cdot f'(r, q-1) ~ + ~ (1 - p_{r}) \cdot f'(r, q) . \notag
\end{align}
We directly see that the definition of $f'(r,q)$ is exactly the same as the corresponding function ``$f(m,r)$'' described in \cite{Bertsimas1992} for the SS-VRP-CD with unit demands, that is, the SS-VRP-C.

\section{Recourse strategy $\mathcal{R}^{q+}$ \label{sec:R_Q_plus} }

\begin{figure}
\centering
\includegraphics[width=1.0\textwidth]{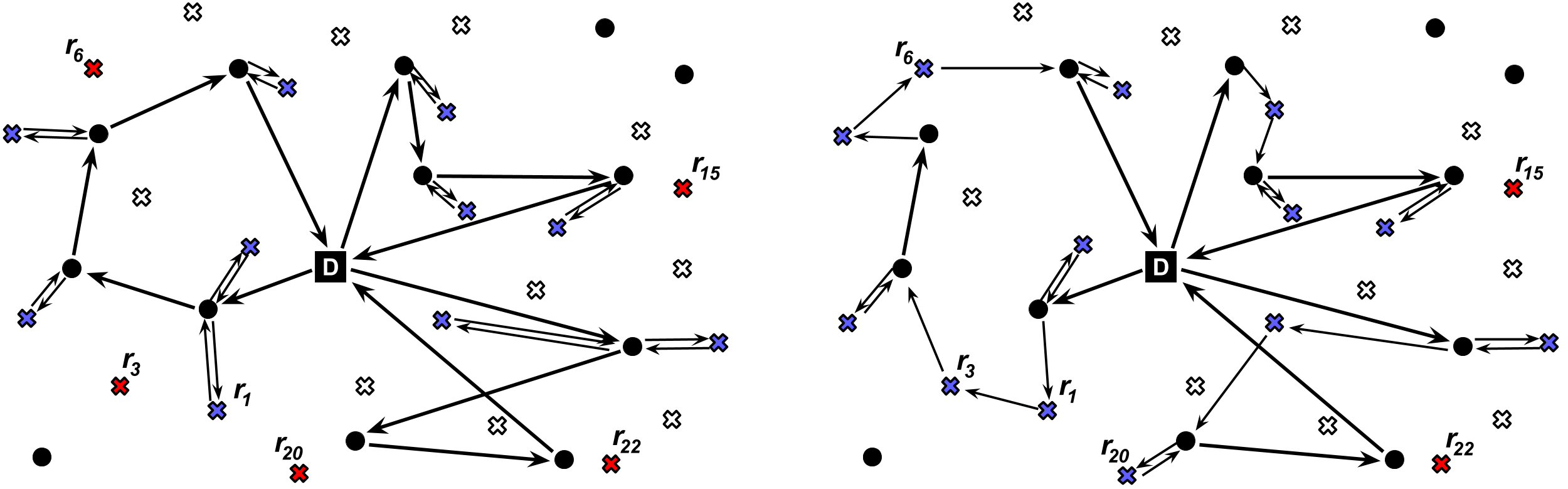}
\caption{
Comparative examples of strategies $\mathcal{R}^q$ (left) and $\mathcal{R}^{q+}$ (right). A filled cross represents a revealed request. 
Under $\mathcal{R}^q$, some requests ($r_3, r_6, r_{15}, r_{20}, r_{22}$) can be missed. 
By avoiding pointless travels when possible, $\mathcal{R}^{q+}$ is likely to end up with a lower number of missed requests than strategy $\mathcal{R}^q$. 
E.g., if request $r_3$ reveals by the time request $r_1$ gets satisfied, then traveling directly to $r_3$ could help satisfying it. 
Similarly in a different route, by traveling directly to the waiting location associated to request $r_{20}$, the vehicle could save enough time to satisfy $r_{20}$.
}
\label{fig:instance_strategies}
\end{figure}

Based on the same potential request assignment and ordering than under $\mathcal{R}^\infty$ and $\mathcal{R}^q$,
strategy $\mathcal{R}^{q+}$ improves by saving operational time.
More specifically, $\mathcal{R}^{q+}$ avoids some pointless round trips from waiting locations, traveling directly towards a revealed request from a previously satisfied one. 
Furthermore, a vehicle is now allowed to travel directly from a customer vertex to a waiting location, without going through the current waiting location. 
Consequently, it can potentially finish a request service later if reaching the next planned waiting location is easier from that request.
Figure \ref{fig:instance_strategies} provides an intuitive illustration of the differences between strategies $\mathcal{R}^q$ and $\mathcal{R}^{q+}$, from an operational point of view.

Under $\mathcal{R}^{q+}$, the location from which the vehicle travels to satisfy a request $r$ can potentially be any customer location $v \in C$, in addition to $\w(r)$.
Given $\w(r)$, let then $t^\text{min+}_{r,v} = \max(\underline{on}(\w(r)), ~ \Gamma_{r}, ~ e_{r} - d_{v,{r}})$ 
be the minimum time at which a vehicle can leave its current location $v \in C \cup \{ \w(r) \}$ in order to satisfy a request $r$.
Again, recall that the request ordering and assignment is the same as under $\mathcal{R}^\infty$ and $\mathcal{R}^q$, that is, based on $t^\text{min}_{r,w}$ and $t^\text{max}_{r,w}$ as described in section \ref{sec:R1_infinite_capacity}.
Therefore, $t^\text{min+}_{r,v}$ will only be useful to request notification and vehicle operations phases.

Let $s(\w(r))$ be the waiting location that follows $w$ in first stage solution $(x,\tau)$.
Since the vehicle is now allowed to travel to $s(\w(r))$ from the customer location $c_r \in C$ of a request $r \in \xi$, we also need a variant of $t^\text{max}_{r,w}$ in order to take the resulting savings into account. We call it $t^\text{max+}_{r,v} = \min \big( l_{r} - d_{v,r} , ~ \underline{on}(\s(\w(r))) - d_{v,r} - s_{r} - d_{r,\s(\w(r))} \big)$. 

\subsubsection*{Vehicle operations}

Whenever a request $r$ must be satisfied under $\mathcal{R}^{q+}$, the vehicle travels to the request, satisfies it during $s_{r}$ time units and then, \emph{instead of traveling back to the depot}, considers the next request $r^\text{next}$, if any, as described in strategy $\mathcal{R}^{\infty}/\mathcal{R}^{q}$ at section \ref{sec:R_inf_recourse}. 
Now, depending on $r^\text{next}$:
\begin{itemize}
\item $r^\text{next}$ does not exist, in which case the vehicle travels back to the depot.
\item $\w(r^\text{next}) = \w(r)$ in which case, if $r^\text{next}$ is already revealed and accepted, the vehicle waits until time $t^\text{min+}_{r^\text{next},r}$ and then 
\emph{travels directly to $r^\text{next}$}, from $r$'s location. 
	Otherwise, $r^\text{next}$ is not known yet ({\em i.e.}, current time $r < \Gamma_{r^\text{next}}$), and the vehicle travels back to waiting location $\w(r)$.
	
\item $\w(r^\text{next}) \ne \w(r)$, in which case the vehicle waits until time $\underline{on}\big(\s(\w(r))\big) - d_{r, \s(\w(r))}$ and then travels to $\s(\w(r))$.
\end{itemize}

\subsubsection*{Request notification}

At a current time $t \ge \Gamma_r$, the time ${\it available}(r)$ at which the vehicle will be able to take care of $r$ is still deterministic and computable.
However, under $\mathcal{R}^{q+}$ a request $r$ can be considered as the vehicle is idle at a customer location $c_{r'} \in C, r' \in \pi_k^w, r' <_R r$, as well as at waiting vertex $w = \w(r)$.
Let $v(r, A^t)$ be the function that returns this location.
Similarly to $y(r, A^t)$, function $v(r, A^t)$ is computable as soon as $t \ge \Gamma_r$.
This is shown in \ref{yv_functions_Rplus}.
The request $r$ is then accepted if and only if:
\begin{align}
y(r, A^t) \le t^\text{max+}_{r,v(r, A^t)} \wedge ~~ q_r + \sum_{r' \in \pi_k \cap A^t} q_{r'}  \le Q .
\end{align} 

\subsection{Expected cost of a second stage solutions under $\mathcal{R}^{q+}$}

Unlike strategy $\mathcal{R}^q$, the satisfiability of a request $r$ not only depends on the current time and vehicle load, but also on the vertex from which the vehicle would leave to serve it.
The candidate vertices are necessarily either the current waiting location $w=\w(r)$ or any vertex hosting one of the previous requests associated to $w$.
Consequently, under $\mathcal{R}^{q+}$ the probability $\text{Pr}\{ r \in A^h \}$ is decomposed over all the possible time, load and vertex configurations:
\begin{align}
\text{Pr}\{ r \in A^h \} &= \sum_{t =t^\text{min+}_{r,w}}^{t^\text{max+}_{r,w}} \sum_{q=0}^{Q-q_r}  g_1^{w}(r, t, q)
		+  \sum_{\substack{r' \in \pi_k^w\\r' <_R r}}  \sum_{t =t^\text{min+}_{r,r'}}^{t^\text{max+}_{r,r'}} \sum_{q=0}^{Q-q_r}   g_1^{r'}(r, t, q) \label{eq:req-satisfiable-R_Q_plus} 
\end{align}
where namely 
\begin{align}
g_1^{w}(r, t, q) \equiv \text{Pr}\{ 	& \text{request $r$ appeared at a time $t' \le t$, the vehicle carries a load of $q$} \notag \\
							& \text{and would serve $r$ from waiting location $w$ at time $t$ if $r$ is accepted} \}  \notag \\
g_1^{r'}(r, t, q) \equiv \text{Pr}\{ 	& \text{request $r$ appeared at a time $t' \le t$, the vehicle carries a load of $q$} \notag \\
							& \text{and would serve $r$ from vertex $c_{r'}$ at time $t$ if $r$ is accepted} \} .
\end{align}
Each tuple $(t, q, v)$ in the summation (\ref{eq:req-satisfiable-R_Q_plus}) represents a feasible configuration for satisfying $r$, and we are interested in the probability $g_1^v(r,t,q)$ that the vehicle is actually available for $r$ while being under one of those states.
The calculus of $g_1^v$ under $\mathcal{R}^{q+}$ is provided in \ref{g1_Rplus}. 
Given $n$ customer locations, $w$ waiting locations, a horizon of length $h$ and vehicle capacity of size $Q$, the computational complexity of computing the whole expected cost $\mathcal{Q}^{\mathcal{R}^{q+}}(x,\tau)$ is in $\mathcal{O}\big(n^2h^3Q \big)$.

\subsubsection*{Space complexity}
A naive implementation of equation (\ref{eq:req-satisfiable-R_Q_plus}) would basically fill up a $n^2 \times h^3 \times Q$ array. 
We draw attention on the fact that even a small instance with $n = Q = 10$ and $h=100$ would then lead to a memory consumption of $10^9$ floating point numbers. Using a common 8-byte representation requires more than 7 gigabytes. 
Like strategy $\mathcal{R}^{q}$, important savings are obtained by noticing that the computation of $g_1$ functions for a given request $r$ under $\mathcal{R}^{q+}$ only relies on the previous request $\prv(r)$. By computing $g_1$ while only keeping in memory the expectations of $\prv(r)$ (instead of all $nh$ potential requests), the memory requirement is reduced by a factor $nh$.

\section{Exact approach for the SS-VRPTW-CR}\label{sec:exact}

Provided a first stage solution $(x,\tau)$ to the SS-VRPTW-CR, sections \ref{sec:R1_infinite_capacity} to \ref{sec:R_Q_plus} describe efficient procedures for computing the objective function $\mathcal{Q}^\mathcal{R}(x,\tau)$, {\em i.e.}, the expected number of rejected requests under a predefined recourse strategy $\mathcal{R}$: recourse strategy ${\cal R}^\infty$ in Section \ref{sec:R1_infinite_capacity}, ${\cal R}^q$ in \ref{sec:R_Q}, and ${\cal R}^{q+}$ in \ref{sec:R_Q_plus}.
In this section, we first provide a stochastic integer programming formulation for the SS-VRPTW-CR.
Then, we describe a branch-and-cut approach that may be used to solve this problem to optimality, {\em i.e.}, find the first stage solution $(x,\tau)$ that minimizes $\mathcal{Q}^\mathcal{R}(x,\tau)$.
The computation of $\mathcal{Q}^\mathcal{R}(x,\tau)$ is from now on considered as a black box.

\subsection{Stochastic Integer Programming formulation \label{sec:sip}}

The problem stated by (\ref{ss-vrptw-general})-(\ref{ss-vrptw-general-2}) refers to a nonlinear stochastic integer program with recourse, which can be modeled as the following simple extended three-index vehicle flow formulation:
\begin{align}
& \underset{x,\tau}{\text{Minimize}} \hspace{1em}  \mathcal{Q}^\mathcal{R}(x,\tau) \label{objective-function} \\
&\text{subject to}& \notag\\
& \hspace{1em} \sum_{j \in W_0} x_{ijk} = \sum_{j \in W_0} x_{jik} = y_{ik}  && \forall ~ i \in W_0, ~k \in [1,K] \label{degree-2} \\
& \hspace{1em} \sum_{k \in [1,K]} y_{0k} \le K && \label{depot-visited} \\
& \hspace{1em} \sum_{k \in [1,K]} y_{ik} \le 1 &&\forall ~ i \in W \label{customer-visited} \\
& \hspace{1em} \sum_{\substack{i \in S \\ j \in W\setminus S}} x_{ijk}  \ge y_{vk} && \forall ~ S \subseteq W, ~ v \in S, ~k \in [1,K] \label{subtour-elim} \\
& \hspace{1em} \sum_{l \in H} \tau_{ilk} = y_{ik}   && \forall ~i \in W, ~ k \in [1,K] \label{tau-variables-one}  \\
& \hspace{1em} \sum_{\substack{i \in W_0 \\ j \in W_0 }} x_{ijk} ~ d_{i,j}  + \sum_{\substack{i \in W \\ l \in H}} \tau_{ilk} ~ l +1 \le h   &&\forall ~ k \in [1,K]  \label{route_length}  \\
& \hspace{1em} y_{ik} \in \{0,1\}  && \forall ~ i \in W_0,  ~ k \in [1,K] \label{y-variables} \\
& \hspace{1em} x_{ijk} \in \{0,1\}  && \forall ~ i,j \in W_0 : ~i \ne j, ~ k \in [1,K]  \label{x-variables} \\
& \hspace{1em}  \tau_{ilk} ~ \in \{0,1\}  && \forall ~ i \in W,  ~ l \in H, ~k \in [1,K]  \label{tau-variables}  
\end{align}
Our formulation uses the following binary decision variables:
\begin{itemize}
\item $y_{ik}~$ equals $1$ iff vertex $i \in W_0$ is visited by vehicle (or route) $k \in [1,K]$;
\item $x_{ijk}~$ equals $1$ iff the arc $(i,j)\in W_0^2$ is part of route $k \in K$; 
\item $\tau_{ilk}~~$ equals $1$ iff vehicle $k$ waits for $1 \le l \le h$ time units at vertex $i$.
\end{itemize}
Whereas variables $y_{ik}$ are only of modeling purposes, yet  $x_{ijk}$ and $\tau_{ilk}$ variables solely define a SS-VRPTW-CR first stage solution.
The sequence $\langle w_1, ..., w_{m'} \rangle_k$ of waiting vertices along any route $k \in [1,K]$ is obtained from $x$. 
By also considering $\tau$, we obtain the {\em a priori} arrival time $\underline{on}(w)$ and departure time $\overline{on}(w)$ at any waiting vertex $w$ in the sequence.
By following the process described in the beginning of section \ref{sec:R1_infinite_capacity}, each  sequence $\pi_k$ is computable directly from $(x,\tau)$.

Constraints (\ref{degree-2}) to (\ref{subtour-elim}) together with (\ref{x-variables}) define the feasible space of the asymmetric Team Orienteering Problem (\citeauthor{Chao1996}, \citeyear{Chao1996}). 
In particular, constraint (\ref{depot-visited}) limits the number of available vehicles.
Constraints (\ref{customer-visited}) ensure that each waiting vertex is visited at most once. 
Subtour elimination constraints (\ref{subtour-elim}) forbid routes that do not include the depot. 
As explained in section \ref{sec:Algo-exact}, constraints (\ref{subtour-elim}) will be generated \emph{on-the-fly} during the search.
Constraint (\ref{tau-variables-one}) ensures that exactly one waiting time $1 \le l \le h$ is selected for each visited vertex. 
Finally, constraint (\ref{route_length}) states that the total duration of each route, starting at time unit 1, cannot exceed $h$.

\paragraph{Symmetries}

The solution space as defined by constraints (\ref{degree-2}) to (\ref{sym}) is unfortunately highly symmetric. 
Not surprisingly, we see that any feasible solution actually reveals to be precisely identical under any permutation of its route indexes $p \in K$. 
In fact, the number of symmetric solutions even grows exponentially with the number of vehicles. 
Provided $K$ vehicles, a feasible solution admits $K! - 1$ symmetries.
In order to remove those symmetries from our original problem formulation, we add the following ordering constraints:
\begin{align}
\sum_{i \in W} 2^i ~ y_{i,p} \le \sum_{i \in W} 2^i ~ y_{i,p+1} ~, \hspace{3em} 1 \le p \le K-1
\label{sym}
\end{align}

\subsection{Branch-and-cut approach  \label{sec:Algo-exact}}
We solve program (\ref{objective-function})-(\ref{sym}) by using the specialized branch-and-cut algorithm introduced by \cite{LaporteLouveaux1993} for tackling stochastic integer programs with complete recourse. This algorithm is referred to as the integer $L$-shaped method, because of its similarity to the $L$-shaped method for continuous problems introduced by \cite{Slyke1969}. 

Roughly speaking, our implementation of the algorithm is quite similar to its previous applications to stochastic VRP's (see e.g. \cite{gendreau1995exact}, \cite{Laporte2002}, \cite{heilporn2011integer}).
As in the standard branch-and-cut scheme, an initial \emph{current problem} (\emph{CP}) is considered by dropping integrality constraints (\ref{x-variables})-(\ref{tau-variables}) as well as the subtour elimination constraints (\ref{subtour-elim}). 
In addition, the $L$-shaped method proposes to replace the nonlinear evaluation function $\mathcal{Q}^\mathcal{R}(x,\tau)$ by a lower bounding variable $\theta$. 
We hence solve a relaxed version of program (\ref{objective-function})-(\ref{tau-variables}), by defining the initial \emph{CP}: 
\begin{align}
(\emph{CP}) \hspace{1em} \underset{x,\tau,z}{\text{Minimize}} \hspace{1em}  z \label{CP}
\end{align}
subject to constraints (\ref{degree-2}), (\ref{depot-visited}), (\ref{customer-visited}), (\ref{tau-variables-one}), (\ref{route_length}), (\ref{sym}) and $z \ge 0$. 
The \emph{CP} is then iteratively modified by introducing integrality conditions throughout the branching process and by generating cuts from constraints (\ref{subtour-elim}) 
whenever a solution violates it. 
These are commonly called \emph{feasibility cuts}. 
Let $(x^\nu, \tau^\nu, z^\nu)$ be an optimal solution to \emph{CP}. 
In addition to feasibility cuts, a lower bounding constraint on $z$, so-called \emph{optimality cut}, is generated whenever a solution comes with a wrong objective value, that is when $z^\nu < \mathcal{Q}^\mathcal{R}(x^\nu,\tau^\nu)$. This way $z$ is gradually tightened upward in the course of the algorithm, approaching the objective value from below.

\begin{figure}[t]
\centering
\includegraphics[width=1\textwidth]{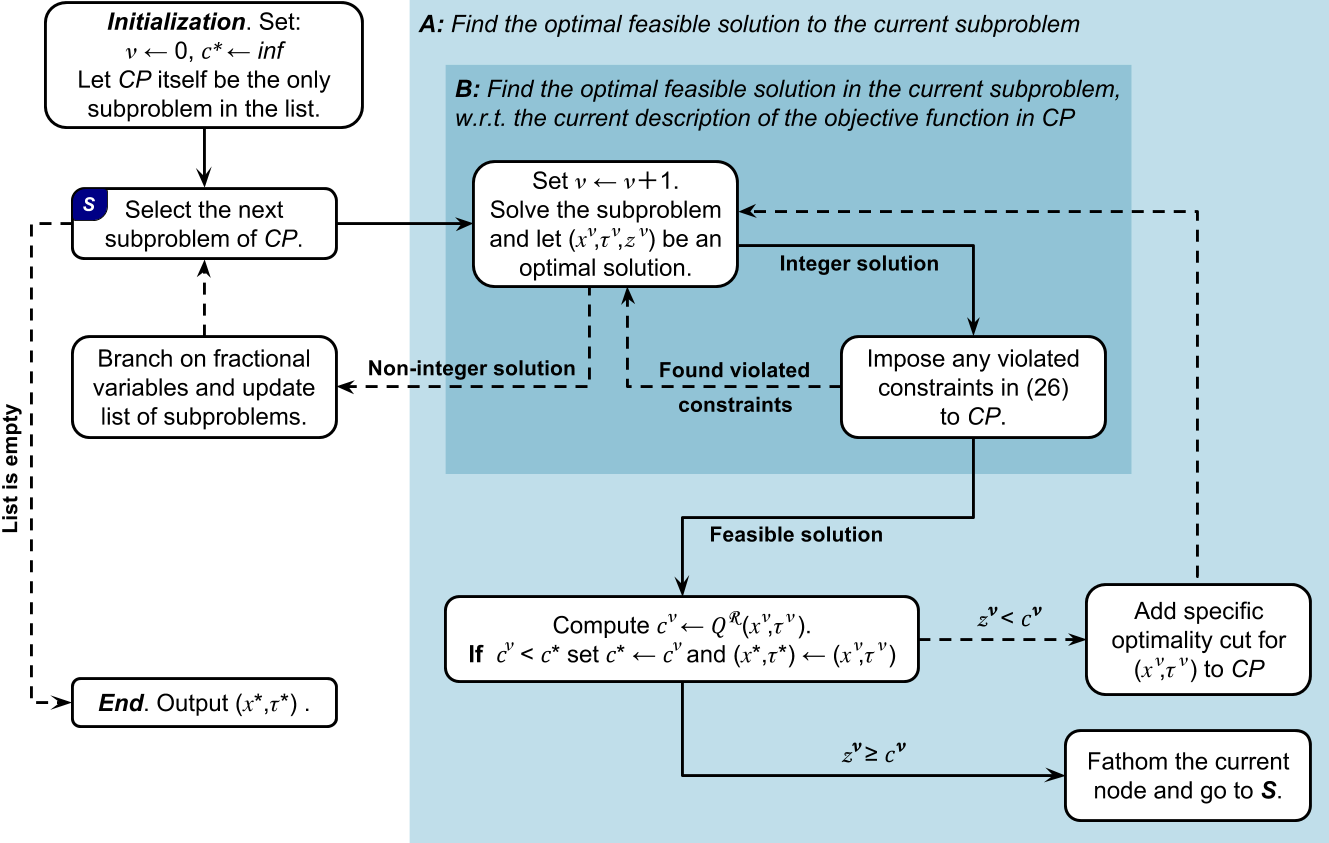}
\caption{
Flow-chart of the branch-and-cut algorithm.
}
\label{fig:algorithm_flow_chart}
\end{figure}

The branch-and-cut scheme is depicted in Figure \ref{fig:algorithm_flow_chart}.
The main steps of the algorithm are as follows. 
Initially, the solution counter $\nu$ and the cost $c^*$ of the best solution found so far are set to zero and infinite, respectively. 
The list of subproblems is initialized in such a way that \emph{CP} is the only problem in it.
Thereafter, the following tasks are repeated until the list of subproblems becomes empty: 1) select a subproblem in the list and 2) find its optimal feasible solution and compare it with the best one found so far. 
When solving a particular subproblem, it is unlikely that \emph{CP} contains the exact representation of the objective function through $z$. Instead, it is approximated by a set of lower bounding constraints. Therefore, whenever an optimal feasible solution $(x^\nu,\tau^\nu, z^\nu)$ is found at the end of frame $B$, it may happen that it is optimal with respect to \emph{CP}, but not with respect to the real objective function. This can be easily checked by comparing the approximated cost $z^\nu$ with the real one $c^\nu = \mathcal{Q}^\mathcal{R}(x^\nu,\tau^\nu)$. If $z^\nu < c^\nu$, then $(x^\nu,\tau^\nu,z^\nu)$ cannot be proven to be optimal since its approximated cost $z^\nu$ is wrong, and consequently better solutions may exist for the current subproblem.

\paragraph{Optimality cuts}

We now present the valid optimality cuts we propose for the SS-VRPTW-CR and that we use in order to gradually strengthen the approximation of $z$ in \emph{CP}. 
Our optimality cuts are  adapted from the classical ones presented in \cite{LaporteLouveaux1993}.
Let $(x^\nu, \tau^\nu, z^\nu)$ be an optimal solution to \emph{CP} where $x^\nu$, $\tau^\nu$ are integer-valued.
Let $A(x^\nu)$ be the set of triples $(i,j,k)$ such that arc $(i,j)$ is part of route $k$, {\em i.e.},
\[A(x^\nu) = \{(i,j,k) : ~ i,j \in W, k \in [1,K] , x_{ijk}^\nu=1\}\]
 and $W(\tau^\nu)$ be the set of triples $(i,l,k)$ such that vehicle $k$ waits $l$ units of time at vertex $i$, {\em i.e.}, 
 \[W(\tau^\nu) = \{(i,l,k) : ~  i \in W, l \in H_0, p \in [1,K], \tau_{ilk}^\nu=1\}.\]
The constraint
\begin{align}
z \ge \mathcal{Q}^\mathcal{R}(x^\nu,\tau^\nu) \Big( \sum_{ \substack{ (i,j,k) \\ \in A(x^\nu)} } x_{ijk} + \sum_{ \substack{ (i,l,k) \\ \in W(\tau^\nu) } } \tau_{ilk} - |A(x^\nu)| - |W(\tau^\nu)| + 1  \Big)  \label{eq:optimality-cut}
\end{align}
is a valid optimality cut for the SS-VRPTW-CR.
Indeed, the integer solution $(x^\nu,\tau^\nu)$ is composed of exactly $|A(x^\nu)|$ arcs and exactly $|W(\tau^\nu)|$ variables in $\tau$ assigned to $1$. For any different solution $(x,\tau,z)$, one must have either $\sum_{(i,j,p) \in A(x)} \le |A(x^\nu)| - 1$ or $\sum_{(i,l,k) \in W(\tau)} \le |W(\tau^\nu)| - 1$. Consequently, for any solution $(x,\tau) \ne (x^\nu, \tau^\nu)$,
\begin{align}
\sum_{(i,j,p) \in A(x)} x_{ijk} + \sum_{(i,l,k) \in W(\tau)} \tau_{ilk} ~ \le ~ |A(x^\nu)| + |W(\tau^\nu)| -1 \label{eq:optimality_cut_2}.
\end{align}
By substituting into (\ref{eq:optimality-cut}), we see that constraint (\ref{eq:optimality-cut}) becomes $z \ge \mathcal{Q}^\mathcal{R}(x^\nu,\tau^\nu)$ only when $(x,\tau) = (x^\nu, \tau^\nu)$. Otherwise the constraint is dominated by $z \ge 0$. 
\hfill $\square$


\section{Local Search for the SS-VRPTW-CR \label{sec:local-search}}

Algorithm \ref{alg:LS-algo} describes a Local Search (LS) approach for approximating the optimal first stage solution $(x,\tau)$, minimizing $\mathcal{Q}^{\mathcal{R}}(x,\tau)$.
The algorithm is the same as the one used in \cite{Saint-Guillain2017}, and it basically follows the Simulated Annealing meta-heuristic of \cite{Kirkpatrick1983}.
The computation of $\mathcal{Q}^{\mathcal{R}}(x,\tau)$ is performed according to equations of sections \ref{sec:R1_infinite_capacity} to \ref{sec:R_Q_plus} (depending on the targeted recourse strategy).
Starting from an initial feasible first stage solution $(x,\tau)$, Algorithm \ref{alg:LS-algo}  iteratively modifies it by using a set of $n_{op}=9$ neighborhood operators.
At each iteration, it randomly chooses a solution $(x',\tau')$ in the current neighborhood (line 4), and either accepts it and resets the neighborhood operator $op$ to the first one (line 5), or rejects it and changes the neighborhood operator $op$ to the next one (line 6).
At the end, the algorithm simply returns the best solution $(x^*,\tau^*)$ encountered so far.
\begin{algorithm}[t]
Let $(x,\tau)$ be an initial feasible first stage solution. \\
Initialize the neighborhood operator $op$ to 1\\
\While{some stopping criterion is not met}{
	
	Select a solution $(x',\tau')$ at random in $\mathcal{N}_{op}(x,\tau)$\\
		
	\If{some acceptance criterion is met on $(x',\tau')$}{
		set $(x,\tau)$ to $(x',\tau')$ and $op$ to 1	}
	\Else{
		change the neighborhood operator $op$ to $op ~ \% ~ n_{op} ~ +1$
	}

}
\textbf{return} the best first stage solution computed during the search
\protect\caption{Local search to compute a first stage solution of SS-VRPTW-CR \label{alg:LS-algo}}
\end{algorithm}

\paragraph{Initial solution and stopping criterion}
The initial first stage solution is constructed by randomly adding each waiting vertex in a route $k \in [1,K]$.
All waiting vertices are thus initially part of the solution.
The stopping criterion depends on the computational time dedicated to the algorithm.

\paragraph{Neighborhood operators}
We consider 4 well known operators for the VRP: relocate, swap, inverted 2-opt, and cross-exchange (see \cite{kindervater1997vehicle,Taillard1997} for detailed description).
In addition, 5 new operators are dedicated to waiting vertices: 2 for either inserting or removing from $W^x$ a waiting vertex $w$ picked at random, 2 for increasing or decreasing the waiting time $\tau_w$ at random vertex $\w \in W^x$, and 1 that transfers a random amount of waiting time units from one waiting vertex to another. 

\paragraph{Acceptance criterion}
We use a Simulated Annealing acceptance criterion. Improving solutions are always accepted, while degrading solutions are accepted with a probability that depends on the degradation and on a temperature parameter, {\em i.e.}, the probability of accepting $(x',\tau')$ is
$e^{- \frac{ 1 - \mathcal{Q}^{\mathcal{R}}(x,\tau) / \mathcal{Q}^{\mathcal{R}}(x',\tau')} { T}}$.
The temperature $T$ is updated by a \emph{cooling factor} $0 < \alpha < 1$ at each iteration of Algorithm \ref{alg:LS-algo}: $T \leftarrow \alpha \cdot T$.
During the search process, $T$ gradually evolves from an initial temperature $T_{\text{init}}$ to  nearly zero. 
A restart strategy is implemented by resetting the temperature to $T \leftarrow T_{\text{init}}$ each time $T$ decreases below a fixed limit  $T_{\text{min}}$.

\section{Experimentations  \label{sec:experimentations}}
In this section, we experimentally compare the $\mathcal{R}^{q}$ and $\mathcal{R}^{q+}$ recourse strategies with a basic "wait-and-serve" policy that does not perform any anticipative actions. The goal is to assess the interest of exploiting stochastic knowledge, and to evaluate the interest of avoiding pointless trips as proposed in $\mathcal{R}^{q+}$ with respect to the basic recourse strategy $\mathcal{R}^{q}$.
We also experimentally compare scale-up properties of the branch-and-cut method and the local search algorithm.

\subsection{A benchmark derived from real-world data \label{sec:instances}}

\begin{figure}[t]
\centering
\includegraphics[width=0.6\textwidth]{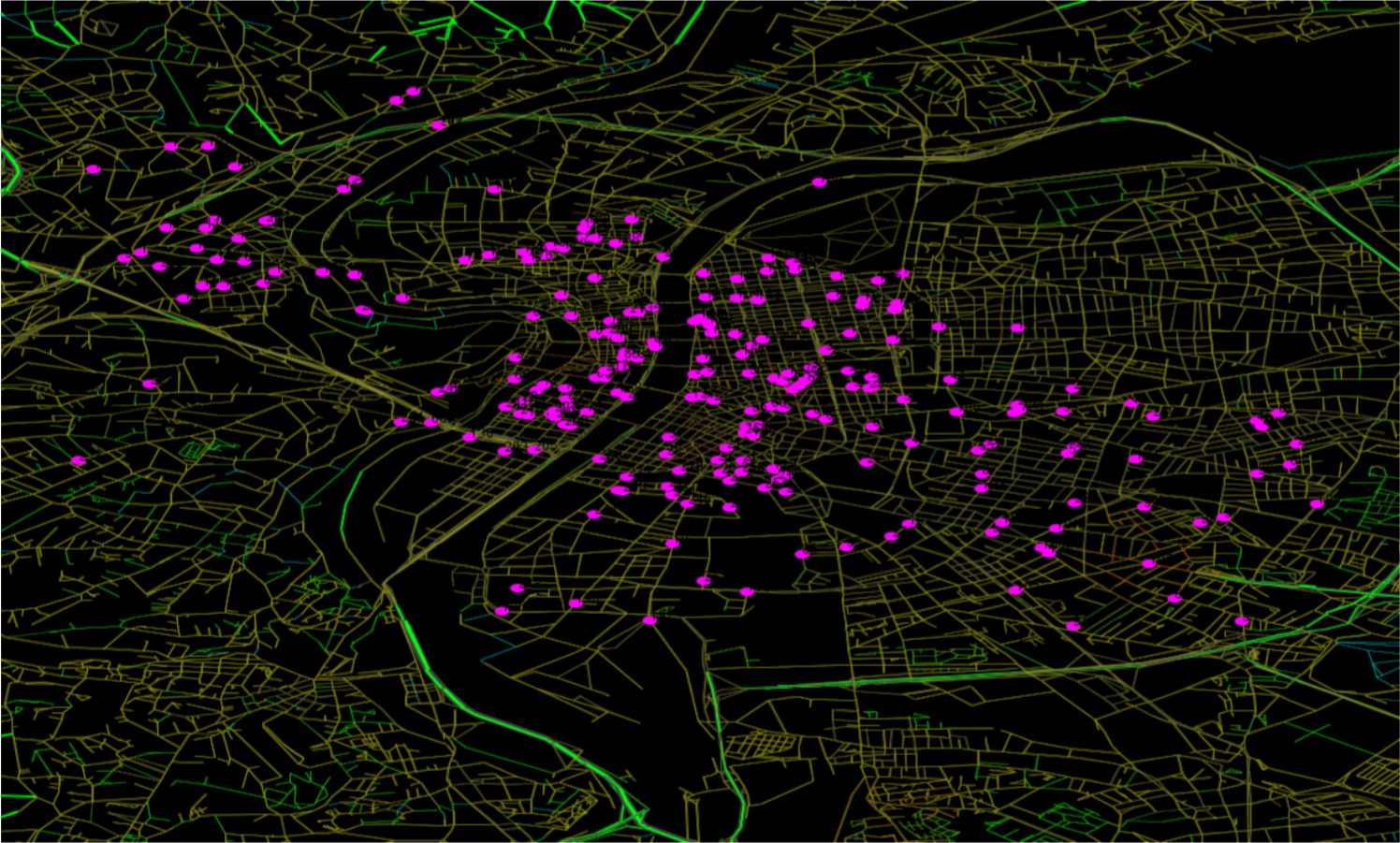}
\caption{
Lyon's road network. In purple, the 255 customer locations.
} \label{fig:lyon}
\end{figure}

\paragraph{Data used to generate instances}
We derive our test instances from the benchmark described in \cite{Melgarejo2015} for the Time-Dependent TSP with Time Windows (TD-TSPTW). This benchmark has been created using real accurate delivery and travel time data coming from the city of Lyon, France.
Travel times have been computed from vehicle speeds that have been measured by 630 sensors over the city (each sensor measures the speed on a road segment every 6 minutes). For road segments without sensors, travel speed has been estimated with respect to speed on the closest road segments of similar type.
Figure \ref{fig:lyon} displays the set of 255 delivery addresses extracted from real delivery data, covering two full months of time-stamped and geo-localized deliveries from three freight carriers operating in Lyon.
For each couple of delivery addresses, travel duration has been computed by searching for a quickest path between the two addresses. In the original benchmark, travel durations are computed for different starting times (by steps of 6 minutes), to take into account the fact that travel durations depend on starting times. In our case, we remove the time-dependent dimension by simply computing average travel times (for all possible starting times). We note $\overline{V}$ the set of 255 delivery addresses, and $d_{i,j}$ the duration for traveling from $i$ to $j$ with $i,j\in \overline{V}$.
This allows us to have realistic travel times between real delivery addresses.
Note that in this real-world context, the resulting travel time matrix is not symmetric. 

\paragraph{Instance generation}
We have generated two different kinds of instances: instances with separated waiting locations, and instances without separated waiting locations.
Each instance with separated waiting locations is denoted $n$\textbf{c}-$m$\textbf{w}-$x$, where $n\in\{10,20,50\}$ is the number of customer locations, $m\in \{5,10,30,50\}$ is the number of waiting vertices, and $x \in [1,15]$ is the random seed.
It is constructed as follows:
\begin{enumerate}
\item We first partition the 255 delivery addresses of $\overline{V}$ in $m$ clusters, using the $k$-means algorithm with $k=m$. During this clustering phase, we have considered symmetric distances, by defining the distance between two points $i$ and $j$ as the minimum duration among $d_{i,j}$ and $d_{j,i}$.
\item For each cluster, we select the median delivery address, {\em i.e.}, the address in the cluster such that its average distance to all other addresses in the cluster is minimal.
The set $W$ of waiting vertices is defined by the set of $m$ median addresses.
\item We randomly and uniformly select the depot and the set $C$ of $n$ customer vertices in the remaining set $\overline{V}\setminus W$.
\end{enumerate}
Each instance without separated waiting locations is denoted $n$\textbf{c}+\textbf{w}-$x$. 
It is constructed by randomly and uniformly selecting the depot and the set $C$ in the entire set $\overline{V}$ and by simply setting $W = C$. 
In other words, in these instances vehicles do not wait at separated waiting vertices, but at customer locations, and every customer location is also a waiting location.

Furthermore, instances sharing the same number of customers $n$ and the same random seed $x$ (\emph{e.g.} 50\textbf{c}-30\textbf{w}-1, 50\textbf{c}-50\textbf{w}-1 and 50\textbf{c+w}-1) always share the exact same set of customer locations $C$.

\paragraph{Operational day, horizon and time slots}
We fix the duration of an operational day to 8 hours in all instances. 
We fix the horizon resolution to $h=480$, which corresponds to one minute time steps.
As it is not realistic to detail request probabilities for each time unit of the horizon (\textit{i.e.}, every minute), we introduce \emph{time slots} of 5 minutes each. 
We thus have $n_{\text{TS}}=96$ time slots over the horizon.
To each \emph{time slot} corresponds a potential request at each customer location. 

\begin{figure}[t]
\centering
\includegraphics[width=1.0\textwidth]{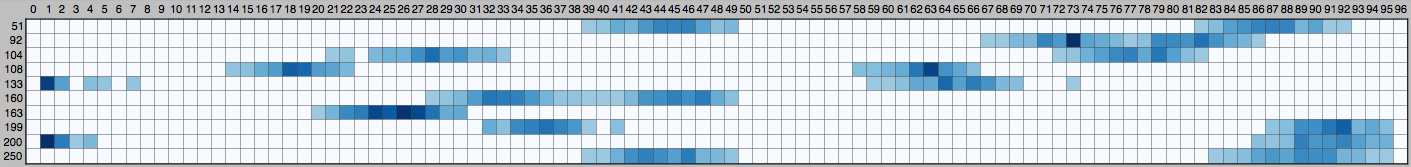}
\caption{
Probability distributions in instance 10c5w-1. 
Each cell represents one of the 96 time slots, for each customer location.
The darker a cell, the likely a request to appear at the corresponding time slot.
A white cell represents a zero probability request that is, no potential request.
}
\label{fig:probas_instance}
\end{figure}

\paragraph{Customer potential requests and attributes.}
For each customer location $c$, we generate the request probabilities associated with $c$ as follows.
First, we randomly and uniformly select two integer values $\mu_1$ and $\mu_2$ in $[1,n_{\text{TS}}]$. 
Then, we randomly generate 200 integer values: 100 with respect to a normal distribution the mean of which is $\mu_1$ and 100 with respect to a normal distribution the mean of which is $\mu_2$. 
Let us note $nb[i]$ the number of times value $i\in [1,n_{\text{TS}}]$ has been generated among the 200 trials. 
Finally, for each reveal time $\Gamma\in H$, if $\Gamma\mod 5 \neq 0$, then we set $p_{(c,\Gamma)}=0$ (as we assume that requests are revealed every 5 minute time slots). 
Otherwise, we set $p_{(c,\Gamma)} = \min(1,\frac{nb[\Gamma/5]}{100})$.
Hence, the expected number of requests at each customer location is smaller than or equal to 2 (in particular, it is smaller than 2 when some of the 200 randomly generated numbers do not belong to the interval $[1,n_{\text{TS}}]$, which may occur when $\mu_1$ or $\mu_2$ are close to the boundary values).
Figure \ref{fig:probas_instance} shows a representation of the distributions in an instance involving 10 customer locations. 

For a same customer location, there may be several requests on the same day at different time slots, and their probabilities are assumed independent.
To each potential request $r=(c_r,\Gamma_r)$ is assigned a deterministic demand $q_r$ taken uniformly in $[0,2]$, a deterministic service duration $s_r=5$ and a time window $[\Gamma_r, \Gamma_r+\Delta-1]$, where $\Delta$ is taken uniformly in $\{5,10,15,20\}$ that is, either 5, 10, 15 or 20 minutes to meet the request. Note that the beginning of the time window of a request $r$ is equal to its reveal time $\Gamma_r$.
This aims at simulating operational contexts similar to the practical application example described in section \ref{sec:Introduction} (the on-demand health care service at home), requiring immediate responses within small time windows.

\paragraph{The benchmark}
The resulting SS-VRPTW-CR benchmark is available at \url{http://becool.info.ucl.ac.be/resources/ss-vrptw-cr-optimod-lyon}. 

\subsection{Experimental plan \label{sec:expe_plan}}
In what follows, experimental concepts and tools we use throughout our experimental analysis are described.
 
\paragraph{Wait-and-serve policy}
In order to assess the contribution of our recourse strategies, we compare them with a case where vehicles do not perform any anticipative operation at all. 
This \emph{wait-and-serve} policy goes as follows.
Vehicles begin the day at the depot. 
Whenever an online request $r$ appears, it is accepted if at least one of the vehicles is able to satisfy it, otherwise it is rejected.
If satisfiable by one of the vehicles, the \emph{closest idle vehicle} satisfies it by directly traveling from its current position to the $r$'s location.
If there are several closest candidates, the least loaded vehicle is chosen.
After the service of $r$ (which lasts $s_r$ time units), the vehicle simply stays idle at $r$'s location until it is assigned another request. 
Eventually, all vehicles must return to the depot before the end of the horizon.
Therefore a request cannot be accepted by a vehicle if serving it prevents the vehicle from returning to the depot in time.

Note that, whereas our recourse strategies for the SS-VRPTW-CR generalize to requests such that the time window starts later than the reveal time (\emph{i.e.}, for which $e_r > \Gamma_r$), in our instances we consider only requests where $e_r = \Gamma_r$.
Doing the other way would in fact require a much more complicated \emph{wait-and-serve} policy, since the current version would be far less efficient (and of poor realism) in case of requests where $e_r$ is significantly greater than $\Gamma_r$.

For the \emph{wait-and-serve} policy, we report average results: we generate $10^6$ scenarios, where each scenario is a subset of $R$ that is randomly generated according to $p_r$ probabilities and, for each scenario, we apply the \emph{wait-and-serve} policy to compute a number of rejected requests; finally, we report the average number of rejected requests for the $10^6$ scenarios.

\paragraph{Scale parameter and waiting time restriction} 
A \emph{scale} parameter is introduced in order to optimize expectations on coarser data, and therefore  speed-up computations.
When equal to 1, expectations are computed while considering the original horizon.
When \emph{scale} $ > 1$, expectations are computed using a coarse version of the initial horizon, scaled down by the factor $scale$.
More precisely, the horizon is scaled to $h_{scale}=h/scale$. 
All the time data, such as travel and service durations, but also time windows and reveal times, are then scaled as well from their original value $v$ to $v/scale$  (rounded up to the nearest integer).
When working on a scaled horizon ({\em i.e.}, scale > 1), algorithms deal with an approximate but easier to compute objective function $\mathcal{Q}^{\mathcal{R}}(x,\tau)$ and a reduced search space due to a coarse time horizon.

Whatever the scale under which an optimization is performed, reported results are always true expected costs, that is expected costs recomputed after scaling solutions up back to the original horizon, multiplying arrival, departure and waiting times by a \emph{scale} factor. 
More precisely, once a first-stage solution is computed, either using the branch-and-cut or the local search method on scaled data, its expected cost is re-expressed according to original data.
This way, we ensure that we can fairly compare results obtained with different scales.

However, even under coarse scales, the problem  is still highly combinatorial. 
In particular, for each waiting location $w$ sequenced in $x$, we need to choose a waiting time $\tau_w$. Even when scaling time, there are still many possible values for $\tau_w$, and the integer programming model described in Section \ref{sec:Algo-exact} associates one decision variable $\tau_{ilk}$ with every waiting vertex $i$, scaled waiting time $l\in [1,h/{\it scale}]$, and vehicle $k\in [1,K]$.
For example, a scale of $2$ leads to a scaled horizon $h/{\it scale}=240$, and therefore $240\cdot mK$ decision variables $\tau_{ilk}$.
Therefore, we limit the possible waiting times of each vehicle at each possible waiting location to a multiple of 10, 30, or 60 minutes. This way, the set of possible waiting times always contains 48, 16 or 8 values (as the time horizon is 8 hours), whatever the scale is. 

\paragraph{Experimental setup} 
All experiments have been done on a cluster composed of 64-bits AMD Opteron 1.4GHz cores.
The code is developed in \emph{C++11}  and compiled with \emph{GCC4.9} and \emph{-O3} optimization flag, using the \emph{CPLEX12.7 Concert} optimization library for the branch-and-cut method.
Current source code of our library for (SS-)VRPs is available from online repository \url{bitbucket.org/mstguillain/vrplib}.
The Simulated Annealing parameters were set to $T_\text{init} = 2, T_\text{min} = 10^{-6}$ and $\alpha = 0.95$.

\subsection{Small instances: Branch\&Cut \emph{versus} Local Search}

\begin{sidewaystable}
\centering
\begin{tabular}{l c r c r    r       c   r         r      c   r         r       c   r               r       c  r               r     c  r                r          }

         & &       & & \multicolumn{8}{c}{Branch\&Cut (\% gain after 12h)}                  &   & \multicolumn{8}{c}{Local Search (\% gain after 1h)}    \\
                           \cmidrule{5-12}                                                      \cmidrule{14-21}                              
         & &       & & \multicolumn{2}{c}{scale = 1} && \multicolumn{2}{c}{scale = 2} && \multicolumn{2}{c}{scale = 5} && \multicolumn{2}{c}{scale = 1} && \multicolumn{2}{c}{scale = 2} && \multicolumn{2}{c}{scale = 5} \\
 \cmidrule{5-6} \cmidrule{8-9} \cmidrule{11-12}             \cmidrule{14-15}     \cmidrule{17-18} \cmidrule{19-21}                          
 && w\&s  && $\mathcal{R}^{q}$     & $\mathcal{R}^{q+}$    && $\mathcal{R}^{q}$       & $\mathcal{R}^{q+}$     && $\mathcal{R}^{q}$       & $\mathcal{R}^{q+}$      && $\mathcal{R}^{q}$      & $\mathcal{R}^{q+}$     && $\mathcal{R}^{q}$     & $\mathcal{R}^{q+}$     && $\mathcal{R}^{q}$      & $\mathcal{R}^{q+}$       \\ \hline
10c-5w-1  && 12.8 && 8.2   & \textbf{14.1}  && 7.3*   & 12.8 * &~& 4.4*   & 9.5*   && 8.2       & 14.0   && 7.3      & 12.8   && 4.4        & 9.5   \\
10c-5w-2  && 10.8 && -4.8  & \textbf{7.4}   && -4.8    & \textbf{7.4}    && -8.8*  & 0.5*   &&  -4.8   &  6.8   &&  -4.8  &  \textbf{7.4}   && -8.8       & 0.5   \\
10c-5w-3  && 8.0  && -46.9 & -26.5 &~& -46.9* & -26.5  && -55.9* & -43.2   && -43.1  & -24.0  && -44.6  & -24.0  && -44.6   &\bf -23.7 \\
10c-5w-4  && 10.5 && -10.9 & \textbf{0.9}   && -10.9* & \textbf{0.9}*  && -10.9* & \textbf{0.9}*   && -10.9   & -6.5   && -11.3  & -4.9   && -10.9     & -3.8  \\
10c-5w-5  && 8.4  && -17.9 & 2.1   &~& -17.9* & \textbf{2.5}    && -20.5* & 1.3*   && -17.9   &  0.9   && -19.5   &  1.0   && -19.5     & 1.3   \\ \hline
\#eval && && $8.10^4$ & $5.10^4$ && $10^5$ & $9.10^4$ && $10^5$ & $9.10^4$ && $2.10^5$ & $10^4$ && $4.10^5$ & $3.10^4$  && $10^6$ & $10^5$ \\ \hline
10c+w-1  && 12.8 && 31.0  & 38.2  && 30.5    & 33.9   && 29.9    & 36.5    &&\bf 40.4    & 36.6   && 39.3    & 38.1   && 34.2    & 33.6  \\
10c+w-2  && 10.8 && 17.9  & 18.8  && 13.2    & 16.9   && 15.2    & 23.6    && 27.9    & 23.1   && 31.6   & 22.6   &&\bf 31.8    & 29.6  \\
10c+w-3  && 8.0  && 12.1  & 16.4  && 14.6    &\bf 32.1   &~& -4.3    & 4.4     && 27.4    & 25.2   && 26.9  & 28.7   && 20.9    & 21.4  \\
10c+w-4  && 10.5 && 8.2   & 12.3  && 10.2    & 14.2   && 3.5     &\bf 28.1    && 22.7   & 19.8   && 23.6   & 23.7   && 18.6   & 22.5  \\
10c+w-5  && 8.4  && 19.7  &\bf 30.4  &~& 5.3     & 22.6   && -0.5    & 20.3    && 27.0  & 24.5   && 28.5   & 26.0   && 24.1   & 20.7  \\ \hline
\#eval && && $5.10^4$ & $3.10^4$ &~& $7.10^4$ & $6.10^4$ && $10^5$ & $9.10^4$ && $2.10^5$ & $10^4$ && $3.10^5$ & $3.10^4$ && $10^6$ & $10^5$ \\ \hline
\end{tabular}
\caption{Results on small instances: $n=10$, $K=2$, $Q = \infty$, and 60 minute waiting time multiples. 
For each instance, we give the average cost over $10^6$ sampled scenarios using the \emph{wait-and-serve} policy (\textit{w\&s}) and the gain of the best solution found by Branch\&Cut within a time limit of 12 hours and Local Search within a time limit of one hour (average on 10 runs). Results marked with a star ($^*$) have been proved optimal.
\#eval gives the average number of expectation computations for each run: optimality cuts (\ref{eq:optimality-cut}) added (Branch\&Cut) or iterations performed (Local Search).
} 
\label{table:results_2veh_infty}
\end{sidewaystable}

Table \ref{table:results_2veh_infty} shows the average gain, in percentages, of using a SS-VRPTW-CR solution instead of the \emph{wait-and-serve} policy on small instances composed of $n=10$ customer locations with $K=2$ uncapacitated vehicles. 
The gain of a solution is $\frac{\text{avg}-E}{\text{avg}}$, where $E$ is the expected cost of the solution and $avg$ is the average cost under the \emph{wait-and-serve} policy.
Unlike the recourse strategies that have to deal with predefined waiting locations, the \emph{wait-and-serve} policy makes direct use of the customer locations.
Therefore, the relative gain of using an optimized first stage SS-VRPTW-CR solution greatly depends on the locations of the waiting vertices. 
Not surprisingly, gains are thus always greater on \emph{10c+w-$x$} instances (as vehicles wait at customer locations): for these instances, gains with the best performing strategy are always greater than $28\%$, whereas for {\em 10c-5w-x} instances, the largest gain is $14\%$, and in some cases gains are negative.

Quite interesting are the results obtained on instance \emph{10c-5w-3}: gains are always negative, {\em i.e.}, waiting strategies always lead to higher expected numbers of rejected requests than the \emph{wait-and-serve} policy. 
By looking further into the average travel times in each instance in Table \ref{table:avg_travel_times_10c}, we find that the travel time between customer locations in instance \emph{10c-5w-3} is rather small (12.5), and very close to time window durations (12.3). In this case, anticipation is of less importance and the \emph{wait-and-serve} policy is performing better.  Furthermore, the travel time between waiting vertices and customer locations (19.5) is much larger than the travel time between customer locations. This penalises SS-VRPTW-CR policies that enforce vehicles to wait on waiting vertices as they need more time to reach customer locations.

\begin{table}[t]
\small
\centering
\begin{tabular}{@{}lrrrrr@{}}
\textit{Instance:}    & 10c-5w-1 & 10c-5w-2 & 10c-5w-3 & 10c-5w-4 & 10c-5w-5 \\ \midrule
\textit{Travel time within $C$:}          & 19.6' & 16.8'  & 12.5' & 18.0' & 13.0'           \\
\textit{Travel time between $C$ and $W$:} & 23.7'   & 19.9'  & 19.5' & 20.9' & 18.2'           \\
\textit{Time window duration:}               & 11.6' & 12.7'  & 12.3' & 13.2' & 12.6'         \\\hline
\end{tabular}
\caption{Statistics on instances 10c-5w-$x$: the first (resp. second) line gives the average travel time between two customer locations of $C$ (resp. between a customer location of $C$ and a waiting vertex of $W$);  and the last line gives the average duration of a time window.
\label{table:avg_travel_times_10c}}
\end{table}

On these small instances, Branch\&Cut runs out of time for all instances when scale=1. 
This is a direct consequence of the enumerative nature of optimality cuts (\ref{eq:optimality-cut}).
Increasing the scale speeds-up the solution process, and allows Branch\&Cut to prove optimality of some \emph{10c-5w-$x$} instances: it proves optimality for 4 and 5 (resp. 2 and 4) instances when scale is set to 2 and 5, respectively, under recourse strategy $\mathcal{R}^{q}$ (resp. $\mathcal{R}^{q+}$). 
In such cases, the computation time needed is around 9 hours under scale 2 and 6.5 hours under $\mathcal{R}^{q}$ at scale 5. 

However, optimizing with coarser scales may degrade solution quality.
This is particularly true for \emph{10c-5w-$x$} instances which are easier than {\em 10c+w-x} instances as they have twice less waiting locations: for these easier instances, gains are often decreased when increasing scales because the search finds optimal or near-optimal solutions, whatever the scale is; however, for harder instances, gains are often increased when increasing scales because the best solutions found when scale=1 are far from being optimal.

For Local Search, gains with recourse strategy $\mathcal{R}^{q+}$ are always greater than gains with recourse strategy $\mathcal{R}^{q}$ on \emph{10c-5w-$x$} instances. 
However, on {\em 10c+w-x} instances, gains with recourse strategy $\mathcal{R}^{q+}$ may be smaller than with recourse strategy $\mathcal{R}^{q}$.
This comes from the fact that computing expectations under strategy $\mathcal{R}^{q+}$ is much more expensive than under strategy $\mathcal{R}^{q}$. Table \ref{table:results_2veh_infty} displays the average number of times the objective function $\mathcal{Q}^{\mathcal{R}}(x,\tau)$ is evaluated (\#eval). For Local Search, this corresponds to the number of iterations of Algorithm \ref{alg:LS-algo}, and this number is around 10 times as small when using $\mathcal{R}^{q+}$ than when using $\mathcal{R}^{q}$, on both \emph{10c-5w-$x$} and {\em 10c+w-x} instances.  \emph{10c-5w-$x$} instances are easier than {\em 10c+w-x} instances (because they have twice as less waiting vertices), and $10^4$ iterations are enough to allow Local Search to find near optimal solutions. In this case, gains with $\mathcal{R}^{q+}$ are much larger than with $\mathcal{R}^{q}$. However, on {\em 10c+w-x} instances, $10^4$ iterations are not enough to find near optimal solutions and, for these instances, better results are obtained with $\mathcal{R}^{q}$ (for which Local Search is able to perform $10^5$ iterations within one hour). 

For Branch\&Cut, the number \#eval of evaluations of the objective function $\mathcal{Q}^{\mathcal{R}}(x,\tau)$ corresponds to the number of times an optimality cut is added. The difference in the number of evaluations when using either $\mathcal{R}^{q}$ or $\mathcal{R}^{q+}$ is less significant (never more than twice as small when using $\mathcal{R}^{q+}$ than when using $\mathcal{R}^{q}$).
Gains with recourse strategy $\mathcal{R}^{q+}$ are always greater than gains with recourse strategy $\mathcal{R}^{q}$, and the improvement is more significant on \emph{10c-5w-$x$} instances (for which waiting locations are different from customer locations) than on {\em 10c+w-x} instances (for which vehicles wait at customer locations).

When optimality has been proven by Branch\&Cut, we note that Local Search often finds solutions with the same gain.
Under scale 2 or 5, Local Search even sometime leads to better solutions: this is due to the fact that optimality is only proven under scale 2 (or 5), whereas the final gain is computed after scaling back to original horizon at scale 1. 
When optimality has not been proven, Local Search often finds better solutions (with larger gains).

In its current basic version, Branch\&Cut is therefore inappropriate even for reasonably sized instances.
We will from now on consider only Local Search for the remaining experiments.

\begin{sidewaystable}
\centering
\begin{tabular}{l c  r    r       c   r         r      c   r         r       c   r           r       c  r         r     c  r          r          }

                & & \multicolumn{8}{c}{Branch\&Cut (\% gain after 12h)}                  &   & \multicolumn{8}{c}{Local Search (\% gain after 1h)}    \\
                           \cmidrule{3-19}                           
               & & \multicolumn{2}{c}{scale = 1} && \multicolumn{2}{c}{scale = 2} && \multicolumn{2}{c}{scale = 5} && \multicolumn{2}{c}{scale = 1} && \multicolumn{2}{c}{scale = 2} && \multicolumn{2}{c}{scale = 5} \\
                           \cmidrule{3-4} \cmidrule{6-7} \cmidrule{9-11}             \cmidrule{12-13}     \cmidrule{15-16} \cmidrule{18-19}                          
Instance && $\mathcal{R}^{q}$     & $\mathcal{R}^{q/q+}$    && $\mathcal{R}^{q}$       & $\mathcal{R}^{q/q+}$     && $\mathcal{R}^{q}$       & $\mathcal{R}^{q/q+}$      && $\mathcal{R}^{q}$      & $\mathcal{R}^{q/q+}$     && $\mathcal{R}^{q}$     & $\mathcal{R}^{q/q+}$     && $\mathcal{R}^{q}$      &  $\mathcal{R}^{q/q+}$       \\ \hline
10c-5w-1  && 8.2   & 9.4  &~& 7.3*   & 12.8*  && 4.4*   &  9.5*   && 8.2    & \textbf{14.1}   && 7.3    &  12.8    && 4.4    & 9.5       \\
10c-5w-2  && -4.8  & \textbf{7.4}  && -4.8    &  \textbf{7.4} && -8.8*  &  0.6*  &&  -4.8  &   \textbf{7.4}  &&  -4.8  &   \textbf{7.4}    && -8.8   & 0.5      \\
10c-5w-3  && -46.9 & -26.5 && -46.9* &  -26.5* && -55.9* &  -37.4*  && -43.1  & \textbf{-22.5}  && -44.6  & -23.7   && -44.6  & -23.7   \\
10c-5w-4  && -10.9 & \textbf{0.9}  && -10.9* & \textbf{0.9}* && -10.9* & \textbf{0.9}*   && -10.9  &   \textbf{0.9}     && -11.3  &   0.3     && -10.9  & \textbf{0.9}      \\
10c-5w-5  && -17.9 &  \textbf{2.5} && -17.9* & \textbf{2.5}*  && -20.5* &  0.5*  && -17.9  &   \textbf{2.5}  && -19.5  &   1.3  && -19.5  & 1.3       \\ \hline
10c+w-1  && 31.0  & 34.2  && 30.5    & 32.6   && 29.9    &  32.3   && 40.4   & \textbf{42.7}   && 39.3   & 41.4     && 34.2   & 35.5     \\
10c+w-2  && 17.9  &  22.0 && 13.2    &  17.2  && 15.2    &   17.7  && 27.9   & 32.9   && 31.6   & \textbf{35.7}     && 31.8   & 34.7     \\
10c+w-3  && 12.1  &  21.6 && 14.6    & 25.1   && -4.3    &  1.9    && 27.4   & \textbf{36.6}   && 26.9   & 36.1     && 20.9   & 28.6     \\
10c+w-4  && 8.2   & 15.7  && 10.2    &  16.1  && 3.5     &  8.6   && 22.7   & 29.7   && 23.6   & \textbf{30.7}     && 18.6   & 24.7     \\
10c+w-5   && 19.7  &  29.7 && 5.3     &  14.1  && -0.5    &  5.2   && 27.0   & 36.3   && 28.5   & \textbf{36.8}     && 24.1   & 29.3    \\ \hline
\end{tabular}
\caption{Comparison of $\mathcal{R}^{q}$ with the hybrid strategy $\mathcal{R}^{q/q+}$ (that uses strategy $\mathcal{R}^{q}$ as evaluation function during the optimization process, and evaluates the final solution with strategy $\mathcal{R}^{q+}$) on the small instances used in Table \ref{table:results_2veh_infty}.} 
\label{table:results_2veh_inftybis}
\end{sidewaystable}

\subsection{Combining recourse strategies}

Results obtained from Table \ref{table:results_2veh_infty} show that although it leads to larger gains, computation of expected costs under recourse strategy $\mathcal{R}^{q+}$ is much more expensive than under $\mathcal{R}^{q}$, which eventually penalizes the optimization process as it performs less iterations within a same time limit (both for Branch\&Cut and for Local Search). 

We now consider a pseudo-strategy we call $\mathcal{R}^{q/q+}$ and that combines $\mathcal{R}^{q}$ and $\mathcal{R}^{q+}$. 
For both Branch\&Cut and Local Search, strategy $\mathcal{R}^{q/q+}$ refers to the process that first uses $\mathcal{R}^{q}$ as evaluation function during the optimization process, and finally evaluates the best final solution using $\mathcal{R}^{q+}$.
Table \ref{table:results_2veh_inftybis} reports the gains obtained by applying $\mathcal{R}^{q/q+}$ on instances \emph{10c-5w-$x$} and \emph{10c+w-$x$}.

For both Branch\&Cut and Local Search, $\mathcal{R}^{q/q+}$ always leads to better results than $\mathcal{R}^{q}$. 
In particular, it also permits Local Search to reach better solutions than using $\mathcal{R}^{q+}$.
This is easily explained by the huge computational effort required by the evaluation function under $\mathcal{R}^{q+}$, which scales down the LS progress. 
By using $\mathcal{R}^{q/q+}$, we actually use $\mathcal{R}^{q}$ to guide the local search, which permits the algorithm to consider a significantly bigger part of the solution space.

From now on, we will only consider Local Search with strategies $\mathcal{R}^{q/q+}$ and $\mathcal{R}^{q+}$ in the next experiments.

\subsection{Impact of waiting time multiples}

Table \ref{table:results_10min_LS} considers instances involving 20 customer locations, and either 10 separated waiting locations (\emph{20c-10w-$x$}) or one waiting location at each customer location (\emph{20c+w-$x$}).
It compares the average results obtained by the LS algorithm when allowing finer-grained waiting times, here multiples of either 60, 30 or 10 minutes, provided either 1 or 12 hours computation time under scale 2.
\begin{sidewaystable}
\centering
\begin{tabular}{l c r r    c    r      r    c   r       r      c    r       r   c     r      r   c    r      r      }
         & & \multicolumn{8}{c}{LS (\% gain after 1h)}            &   & \multicolumn{8}{c}{LS (\% gain after 12h)} \\
         \cmidrule{3-10}                                                 \cmidrule{12-19}
\textit{Wait. mult.:}   & & \multicolumn{2}{c}{60'} & & \multicolumn{2}{c}{30'} & & \multicolumn{2}{c}{10'} & & \multicolumn{2}{c}{60'} & & \multicolumn{2}{c}{30'} & & \multicolumn{2}{c}{10'} \\
              \cmidrule{3-4}   \cmidrule{6-7}  \cmidrule{9-10}         \cmidrule{12-13}  \cmidrule{15-16} \cmidrule{18-19}
  &&  $\mathcal{R}^{q/q+}$ & $\mathcal{R}^{q+}$ && $\mathcal{R}^{q/q+}$ & $\mathcal{R}^{q+}$ && $\mathcal{R}^{q/q+}$ & $\mathcal{R}^{q+}$ && $\mathcal{R}^{q/q+}$ & $\mathcal{R}^{q+}$ &&  $\mathcal{R}^{q/q+}$ & $\mathcal{R}^{q+}$ && $\mathcal{R}^{q/q+}$ & $\mathcal{R}^{q+}$  \\ \hline
20-c10w-1 && 9.6  & 4.6  && 11.3  &  5.2 && \textbf{14.5}  &  3.9   && 11.0  &  8.6  && 12.0  & 10.8 && \textbf{15.3} & 14.1  \\
20-c10w-2 && -5.7 & -13.8&& -4.3  &-11.9 && \textbf{-2.8}  & -8.6   && -5.7  & -8.4  && -4.8  & -4.8 && \textbf{-2.4} & -2.7  \\
20-c10w-3 && 1.2  & -2.6 &&  7.5  & -2.7 &&  \textbf{7.9}  & -3.1   &&  1.1  &  0.2  &&  7.8  &  7.3 && \textbf{8.1}  & 5.8   \\
20-c10w-4 && 5.5  & 4.1  &&  \textbf{5.6}  &  4.6 &&  5.4  &  3.8   &&  5.5  &  \textbf{6.1}  &&  4.8  &  5.7 && 4.7  & 5.9   \\
20-c10w-5 && -10.5& -13.4&&  \textbf{0.0}  & -1.9 && -0.3  & -7.8   && -8.9  & -9.5  && -0.5  &  0.2 && \textbf{0.4}  & -0.2  \\
\hline
20-c+w-1  && 15.9 & 12.1 && 17.3  &  8.5 && \textbf{17.8}  & 11.9   && 18.7  & 15.1  && 21.3  & 16.5 && \textbf{22.9} & 17.9  \\
20-c+w-2  && \textbf{8.2}  & -2.5 &&  6.1  &  0.2 &&  7.7  &  0.6   && \textbf{14.1}  &  7.0  && 11.7  &  5.0 && 14.0 & 8.0  \\
20-c+w-3  && 3.0  & -1.1 &&  5.2  & -1.1 &&  \textbf{6.8}  & -0.4   &&  7.2  &  3.7  &&  7.9  &  3.8 && \textbf{10.5} & 4.3  \\
20-c+w-4  && \textbf{17.2} & 11.7 && 16.6  & 13.5 && 14.4  & 11.8   && 18.8  & 16.6  && \textbf{20.2}  & 16.0 && 20.1 & 18.1  \\
20-c+w-5  && 12.5 & 6.4  && 11.7  &  6.8 && \textbf{17.6}  & 11.5   && 18.6  &  9.6  && 17.6  & 14.3 && \textbf{18.9} & 17.2  \\ 
\hline
\#eval && $15.10^4$ & $6.10^3$ &~& $15.10^4$ & $6.10^3$ && $15.10^4$ & $6.10^3$ && $2.10^6$ & $8.10^4$ && $2.10^6$ & $8.10^4$ && $2.10^6$ & $8.10^4$ \\ 
\hline
\end{tabular}
\caption{Relative gains when varying computation times and the granularity of waiting time multiples, using $K=2$ uncapacitated vehicles under scale = 2.
\#eval gives the average number of iterations performed by Local Search, which in average does not depends on the number $|W|$ available waiting locations.} 
\label{table:results_10min_LS}
\end{sidewaystable}

Despite the increasing complexity of dealing with 20 customer locations and 10 waiting locations, finer-grained waiting time multiples of 30 and 10 minutes tend to provide better solutions than 60 minutes.
Also note that the LS approach still tends to provide promising average gains compared to the \emph{wait-and-serve} policy, in particular when waiting locations coincide with customer locations (\emph{10c+w-$x$} instances). 

Finally, over 12 hours of computation the difference in the number of iterations performed under either $\mathcal{R}^{q}$ or $\mathcal{R}^{q+}$ is now around 25 times smaller in the later case. 
Recall that with instances \emph{10c-5w-$x$} and \emph{10c+w-$x$} (\textit{i.e.} when $n=|C|=10$), the difference was of only 10 times. Hence, when limiting the CPU time to 1 hour, results obtained with $\mathcal{R}^{q/q+}$ are always better than with $\mathcal{R}^{q+}$. However, when increasing the CPU time limit to 12 hours, the difference between the two strategies decreases and in some cases (such as instance \emph{20c-10w-4}, for example), $\mathcal{R}^{q+}$ outperforms $\mathcal{R}^{q/q+}$.

\subsection{Local Search variants on bigger instances \label{sec:big_instances}}
We now consider the following instance classes: \emph{50c-30w-$x$}, \emph{50c-50w-$x$} and \emph{50c+w-$x$}, each with either 5, 10 and 20 vehicles. 
In all cases, the vehicle's capacity is now constrained to $Q=20$.
Each class is composed of 15 instances. The three classes are such that, for each seed $x\in [1,15]$, the three instances \emph{50c-30w-x}, \emph{50c-50w-x} and \emph{50c+w-x} contain the same set of 50 customer locations, and thus only differ on the number and/or position of the available waiting locations.

\paragraph{Considered Local Search variants}
We compare seven different variants, that are all run with a same CPU time limit of twelve hours. Also, for all variants, the expected cost is computed under strategy $\mathcal{R}^{q}$ during the whole twelve hour LS process; the expected cost of the final solution is finally evaluated under scale 1 using strategy $\mathcal{R}^{q+}$.

The first four variants correspond to different instantiations of the LS algorithm with respect to scale and waiting time parameters: we consider two different scales $a\in \{1,2\}$, and two different waiting time multiples $b\in \{10,60\}$, and denote \emph{LS-s$a$-w$b$} the LS algorithm using scale $a$ and waiting time multiples of $b$ minutes.

We also consider three approaches that change some parameter settings during the LS solving process:
\begin{enumerate}[$\diamond$~~]
\item \emph{LS-s*-w10}, where waiting time multiples are set to 10 minutes, for the whole twelve hour solving process, while the scale is progressively decreased every four hours from 5, to 2, and finally 1;
\item \emph{LS-s1-w*}, where scale is set to 1, for the whole twelve hour solving process, while waiting time multiples are progressively decreased every four hours from 60, to 30, and finally 10 minutes;
\item \emph{LS-s*-w*}, where both scale and waiting time multiples are progressively decreased. Waiting time multiples are decreased every four hours from 60, to 30, and finally 10 minutes. Within each of these four hour steps, scale is set to 5 for the first 80 minutes; it is then decreased to 2 for the next 80 minutes and finally set to 1 for the last 80 minutes.
\end{enumerate}

\begin{figure}
\centering
\begin{minipage}[c]{.47\linewidth}
      \includegraphics[width=1.0\textwidth]{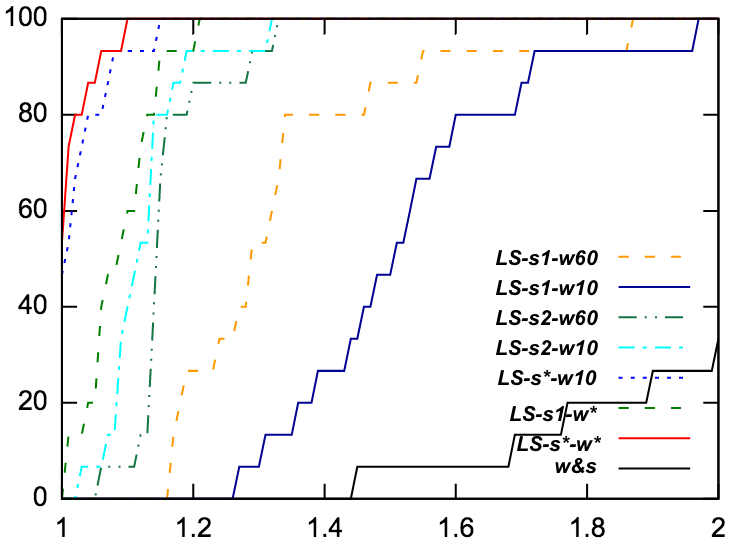}
   \end{minipage} \hfill
   \begin{minipage}[c]{.47\linewidth}
      \includegraphics[width=1.0\textwidth]{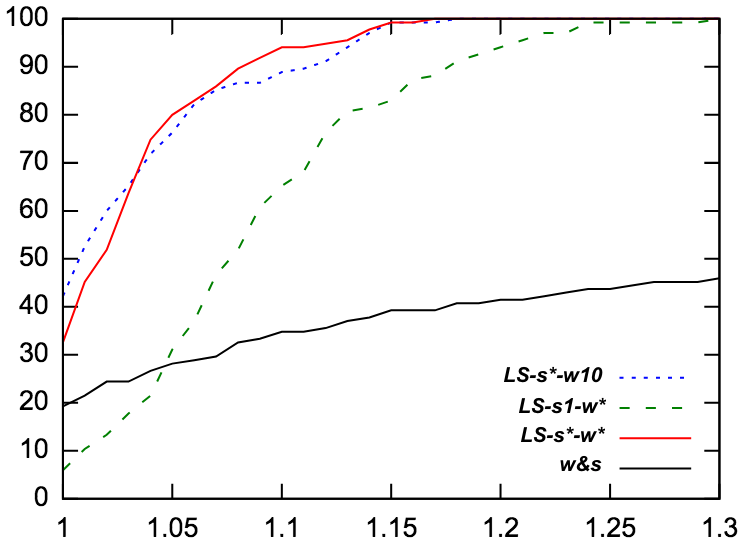}
   \end{minipage}
\caption{
Performance profiles.
Left: comparing all seven LS variants and the \emph{w\&s} policy on the 15 instances of class \emph{50c+w-$x$} only, using 20 vehicles capacitated vehicles.
Right: comparing \emph{LS-s*-w10}, \emph{LS-s10-w*} and \emph{LS-s*-w*} on the 3 classes (\emph{50c-30w-$x$}, \emph{50c-50w-$x$}, \emph{50c+w-$x$}), with \{5, 10, 20\} capacitated vehicles (135 instances in total).
}
\label{fig:performance_profiles}
\end{figure}

The performance of these seven LS variants and the baseline \emph{wait-and-serve} policy  are compared by using so-called \emph{performance profiles}.
This is depicted on Figure \ref{fig:performance_profiles}.
Performance profiles provide, for each considered approach, a cumulative distribution of its performance compared to other approaches. 
For a given approach A, a point $(x,y)$ on A's curve means that in $(100\cdot y)\%$ of the instances, A performed at most $x$ times worse than the best approach on each instance taken separately. 
An approach A is strictly better than another approach B if A's curve stays always above B's curve.
See \cite{dolan2002benchmarking} for a complete description of performance profiles. 

According to Figure \ref{fig:performance_profiles} (left), algorithms \emph{LS-s*-w10} and \emph{LS-s*-w*} show the best performances (which look quite similar) when tested on instance class \emph{50c+w-$x$} only. 
More experiments are conducted to distinguish between algorithms \emph{LS-s*-w10}, \emph{LS-s1-w*} and \emph{LS-s*-w*}, on all three instance classes and using 5, 10 and 20 capacitated vehicles. 
This is shown on Figure \ref{fig:performance_profiles} (right). 

In comparison to the other approaches, algorithms \emph{LS-s*-w10} and \emph{LS-s*-w*} clearly obtain the best performances in average. 
In what follows, we use algorithm \emph{LS-s*-w*} only in order to analyze the influence of operational settings (\emph{i.e.} number of vehicles, waiting locations, time windows) on the average gains.

\subsection{Influence of the number of vehicles \label{sec:big_instances_number_veh}}
\begin{figure}
\centering
\includegraphics[width=1.0\textwidth]{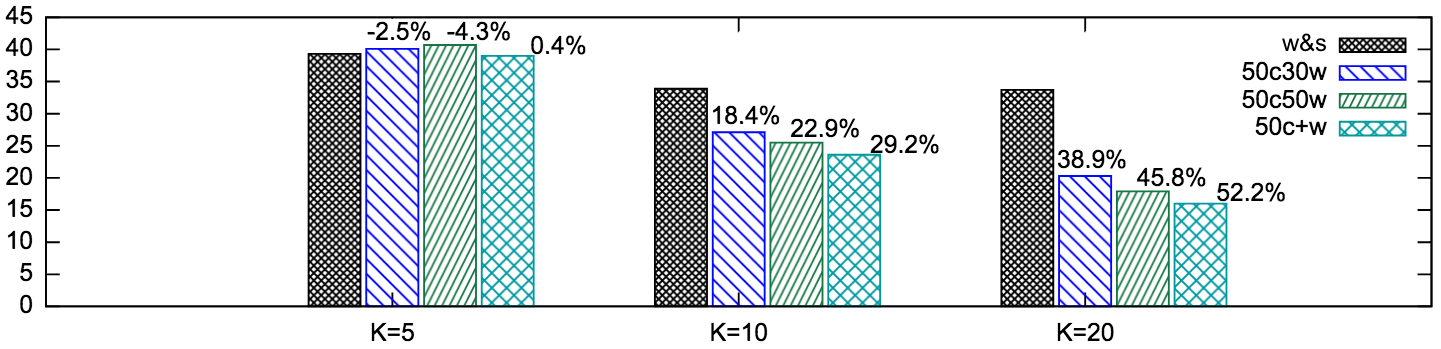}
\caption{
Average number of rejected requests on instance classes \emph{50c-30w-$x$}, \emph{50c-50w-$x$}, \emph{50c+w-$x$}, while varying the number $K$ of vehicles.
The height of each bar gives the average (\emph{resp.} expected) number of rejected requests of the \emph{w\&s} policy (\emph{resp.} SS-VRPTW-CR second stage solution). 
}
\label{fig:hist_tw_x1}
\end{figure}

Figure \ref{fig:hist_tw_x1} shows how the performance of the SS-VRPTW-CR model relatively to the \emph{wait-and-serve} policy varies with the waiting locations and the number of vehicles. For 5, 10 and 20 vehicles, the average over each of the instance classes (15 instances per class) is reported. 
Solution files and detailed result tables are available at \url{http://becool.info.ucl.ac.be/resources/ss-vrptw-cr-optimod-lyon}.

It shows us that the more vehicles are involved, the more important are clever anticipative decisions and therefore, the more beneficial is a SS-VRPTW-CR solution compared to the \emph{wait-and-serve} policy.
It is likely that, as conjectured in \cite{Saint-Guillain2017}, a higher number of vehicles leads to a less uniform objective function, with most probably steepest local optima. 
Because it requires much more anticipation than given only 5 vehicles, using the SS-VRPTW-CR model instead of the \emph{wait-and-serve} policy reveals to be particularly beneficial provided (more than) 10 or 20 vehicles. With 20 vehicles, our model decreases the average number of rejected request by 52.2\% when vehicles are allowed to wait at customer locations (\emph{i.e.}, on instance class \emph{50c+w-$x$}). 

\subsection{Influence of the time windows}
\begin{figure}
\centering
\includegraphics[width=1.0\textwidth]{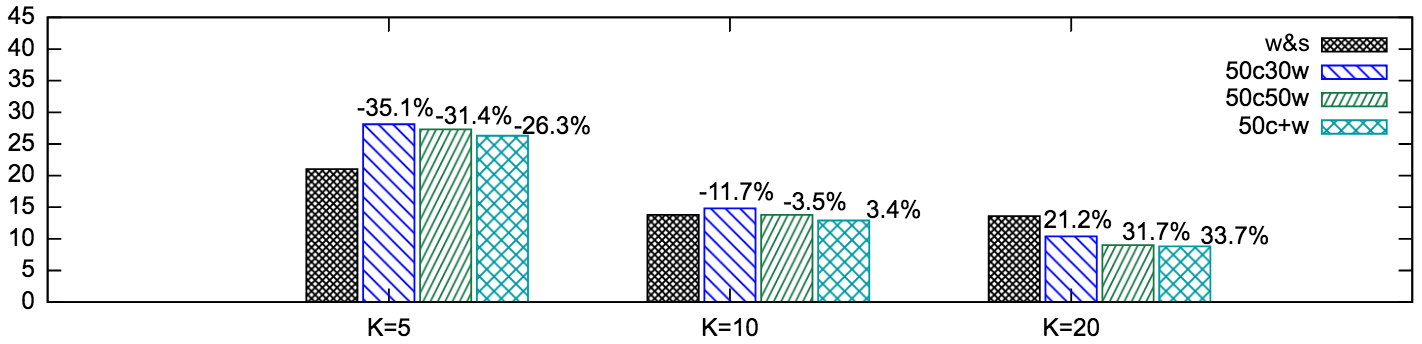}
\includegraphics[width=1.0\textwidth]{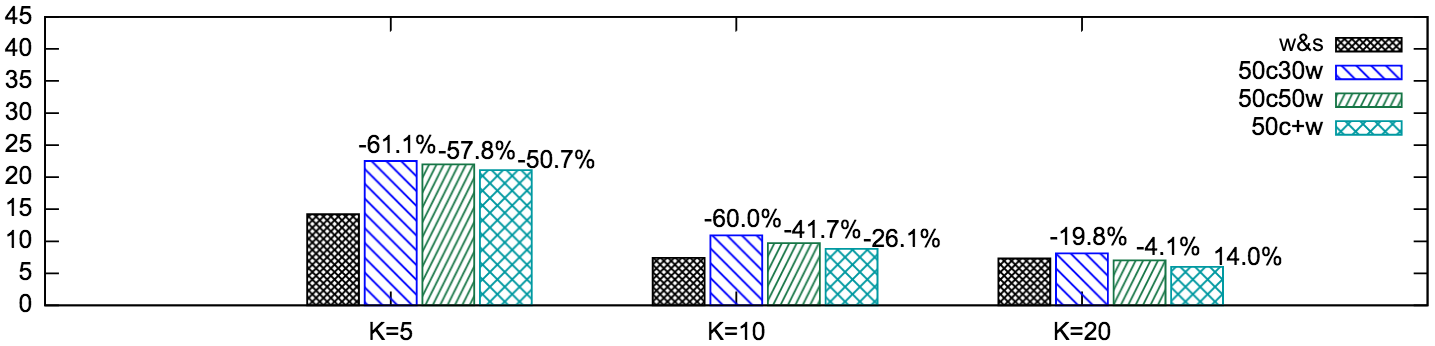}
\caption{
Average number of rejected requests when all time window durations are doubled (top) and tripled (down).
}
\label{fig:hist_tw_x2_x3}
\end{figure}

We now consider less demanding customers by conducting the same experiments as in section \ref{sec:big_instances_number_veh}, while modifying potential request time windows only.
Top of Figure \ref{fig:hist_tw_x2_x3} shows the average gain of using a SS-VRPTW-CR model when the service quality is lowered down by \emph{multiplying all the original time window durations by two}. 

The results show that for a $K=5$ vehicles, the \emph{wait-and-serve} policy always performs better. 
With $K=20$ vehicles however, the average relative gain of using the SS-VRPTW-CR model stays significant: 33.7\% less rejected requests in average on instance class \emph{50c+w-$x$}. 

Bottom of Figure \ref{fig:hist_tw_x2_x3} illustrates how the average gain evolves when \emph{multiplying all the original time windows by three}. 
Provided 20 vehicles, the SS-VRPTW-CR model still improves the \emph{wait-and-serve} policy by 14\% when vehicles are allowed to wait directly at customer locations. 
Together with Figure \ref{fig:hist_tw_x1}, Figure \ref{fig:hist_tw_x2_x3} shows that the SS-VRPTW-CR model is more beneficial when the number of vehicles is high and the time windows are small.

\subsection{Position of the waiting locations}
From all the experimentations that have been conducted, what immediately appears is that no matter the operational context into consideration (the number of customer locations, the number of vehicles) or the approximations that simplify the problem (scaling factor, waiting time multiples), \emph{allowing the vehicles to wait directly at customer locations always lead to better results than using separated waiting vertices}. 
Except if one has constraints limiting the possible waiting locations (\emph{e.g.,} if vehicles cannot park for too long anywhere in the city), placing waiting vertices in such a way that they coincide with customer locations is always the best choice.

\section{Conclusions and research directions\label{sec:conclusions}}
In this paper, the SS-VRPTW-CR is formally defined together with some of its possible recourse strategies. 
We extend the model previously introduced in \cite{Saint-Guillain2017} for uncapacitated vehicles, by presenting two additional recourse strategies: $\mathcal{R}^{q}$ and $\mathcal{R}^{q+}$. 
These take customer demands into account and allow the vehicles to save operational time, traveling directly between customer locations when possible.
We show how, under the later recourse strategies, the expected cost of a second stage solution is computable in pseudo-polynomial time.

Proof of concept experiments on small and reasonably large test instances compare these anticipative models with each other and show their superiority compared to a basic "wait-and-serve" policy. 
These preliminary results confirm that, although computationally more demanding, optimal first stage solutions under $\mathcal{R}^{q+}$ generally lead to significantly lower expected cost than under $\mathcal{R}^{q}$. 

The Local Search method proposed in \cite{Saint-Guillain2017} produces near-optimal solutions on small instances.
In this paper, we also present improved variants of that algorithm, and demonstrate their efficiency on bigger instances.
This is done by exploiting heuristics and scaling techniques. 
We show that our heuristic methods eventually allows to tackle larger problems, for which the exact approach is not possible. 

We show that SS-VRPTW-CR recourse strategies provide significant benefits compared to a non-anticipative (but yet realistic) policy.
Results on a variety of large instances also show that the benefit of using the SS-VRPTW-CR increases with the number of vehicles involved and the level of responsiveness imposed by the time windows.
Finally, all our experiments indicate that allowing the vehicles to wait directly at potential customer locations lead to better expected results than using separated relocation vertices. 

\subsection*{Future work}

\paragraph{On solving methods}
The branch-and-cut approach considered in this paper may not be perfectly designed for the SS-VRPTW-CR. 
First notice that each optimality cut of the form (\ref{eq:optimality-cut}) is likely to be active at only one feasible solution. 
As pointed out by \cite{Hjorring1999}, if only these cuts are to be added to \emph{CP}, then our branch-and-cut method must generate such a cut for almost each feasible first stage solution.  
Thus, although not trivially doable, general optimality cuts that are active at fractional solutions should be devised as well.
Furthermore, branch-and-cut is only one method among all the possible approaches that could be tested on the SS-VRPTW-CR. 
In particular set partitioning methods such as column generation, which becomes commonly used for stochastic VRPs, could also provide interesting results. 

\paragraph{On scaling techniques}
We have shown through experiments that the computational complexity of the evaluation of the objective function in SS-VRPTW-CR is an issue that can be successfully addressed by scaling down the problem instances. 
However, the scale is always performed in terms of temporal data, as a consequence of the reduction of the horizon. 
It may be interesting to also consider scaling the set of potential requests, such as geographical clustering for instance, that would also allow us to significantly reduce the computational effort when evaluating a first stage solution.

\paragraph{Further application to online optimization}
As already pointed out in \cite{Saint-Guillain2017},
another potential application of the SS-VRPTW-CR goes to online optimization problems such as the Dynamic and Stochastic VRPTW (DS-VRPTW). 
Most of the approaches that have been proposed in order to solve the DS-VRPTW rely on reoptimization. 
However, because of the intractable computational complexity involved in exact online reoptimization, heuristic methods are often preferred. 
The most common approximation is called Sample Average Approximation (SAA, see \citeauthor{ahmed2002sample}, \citeyear{ahmed2002sample} and \citeauthor{verweij2003sample}, \citeyear{verweij2003sample}), and relies on Monte Carlo sampling to evaluate only a subset of the possible scenarios on a current first stage solution. 
As a consequence, the approximate cost associated to the first stage solution critically depends on the number of samples considered, and does not constitute an upper bound on the true expected cost under optimal reoptimization.

Thanks to recourse strategies, the SS-VRPTW-CR finally provides an actual guarantee on expected costs: it provides an upper bound on the expected cost of a first stage solution under optimal reoptimization. 
The SS-VRPTW-CR can therefore be used as a subroutine in order to heuristically solve the DS-VRTPW, whilst considering the whole set of scenarios instead of only a limited subset of sampled ones.

\paragraph{Towards better recourse strategies}
The expected cost of a first stage solution obviously depends on how the recourse strategy fits the operational problem.
Improving these strategies may tremendously improve the quality of the upper bound they provide to exact reoptimization.
The recourse strategies presented in this paper are of limited operational complexity, yet the associated computational complexity is already really expensive. 
One potential improvement, which limits the increase in computational requirements, may be to rethink the way the potential requests are assigned to waiting locations (\emph{e.g.} by taking their probabilities and demands into account as well).  
Another direction would be to think at better, more intelligent, vehicle operations.
However, a somehow important question remains: how intelligent could be a recourse strategy so as its expected cost stays efficiently computable?

\subsection*{Acknowledgements}
Computational resources have been provided by the Consortium des Équipements de Calcul Intensif (CÉCI), funded by the Fonds de la Recherche Scientifique de Belgique (F.R.S.-FNRS) under Grant No. 2.5020.11.
Christine Solnon is supported by the LABEX IMU (ANR-10-LABX-0088) of Université de Lyon, within the program "Investissements d’Avenir" (ANR-11-IDEX-0007) operated by the French National Research Agency (ANR).

\bibliography{library-noURLs.bib}

\appendix

\section{Computing $y(r)$ and $v(r)$ under $\mathcal{R}^{q+}$ \label{yv_functions_Rplus}}

Given current time $t \ge \Gamma_r$ and previous request $r' \in \prv(r)$, function $y(r)$ must be adapted to strategy $\mathcal{R}^{q+}$:
\begin{align}
y(r) 	&= \begin{cases}
		\max \big( y(r')+ d_{v(r'), r'}, e_{r'} \big) + s_{r'} + d_{r', v(r)}, & \text{if } ~~ r' \in A^t  \\
		y(r') + d_{v(r'), v(r)}, & \text{otherwise}
	\end{cases}  \notag
\end{align}
with 
$$y(r_1) = \underline{on}(w), ~r_1 = \fst(\pi_k^w), w = \w(r_1) = \w(r) .$$
The location $v(r)$ from which the vehicle travels towards request $r$ depends on whether $r$ reveals by the time the vehicle finishes to satisfy the last accepted request: 
\begin{align}
v(r) 	&= \begin{cases}
		c_{r'}, 		& \text{if } ~~ \Gamma_r \le  \max \big( y(r')+ d_{v(r'), r'}, e_{r'} \big) + s_{r'}  ~~ \wedge ~~ r' \in A^t \\
		v(r'), 	& \text{if } ~~ r' \notin A^t   \\
		\w(r)  		& \text{otherwise}
	\end{cases}  \notag
\end{align}
with base case $v(r_1) = \w(r_1) = \w(r)$, $r_1 = \fst(\pi_k^w)$.

\section{Computing $g_1^v(r,t,q)$ under strategy $\mathcal{R}^{q+}$ \label{g1_Rplus}}

Let the following random functions:
\begin{align}
h^w(r,t,q) \equiv \text{Pr}\{ 	& \text{the vehicle gets rid of request $r$ at time $t$ with a load of $q$} \notag \\ 
							& \text{and at waiting location $w$} \} \notag \\
h^{r'}(r,t,q) \equiv \text{Pr}\{ 	& \text{the vehicle gets rid of request $r$ at time $t$ with a load of $q$} \notag \\ 
							& \text{and at location $c_{r'}$} \} \notag
\end{align}
and
\begin{align}
g_2^w(r, t, q) \equiv \text{Pr}\{ 	& \text{request $r$ did not appear and the vehicle discards it at time $t$} \notag \\ 
							& \text{with a load of $q$ while being at waiting location $w$} \} \notag \\
g_2^{r'}(r, t, q) \equiv \text{Pr}\{ 	& \text{request $r$ did not appear and the vehicle discards it at time $t$} \notag \\ 
							& \text{with a load of $q$ while being at location $c_{r'}$} \}. \notag
\end{align}
For the very first request $r_1^k = \fst(\pi_k)$ of the route, trivially the current load $q$ of the vehicle must be zero, and it seems normal for the waiting location $w = \w(r_1^k)$ to be the only possible location from which the vehicle can be available to handling $r_1^k$ if the request appears, or to discard it if it doesn't:
\begin{align}
g_1^w(r_1^k,t,q) = 	 \begin{cases}
	p_{r_1^k}, & \text{if } ~~ t = t^\text{min+}_{{r_1^k},w} ~~\wedge~~ q = 0 \\ 
	0  		& \text{otherwise}.
	\end{cases} 		\notag	
\end{align}
\begin{align}
g_2^w(r_1^k,t,q) =   \begin{cases}
	1 - p_{r_1^k}, & \text{if } ~~ t = \max( \underline{on}(w), \Gamma_{r_1^k}) ~~\wedge~~ q = 0 \\
	0  & \text{otherwise}.
	\end{cases} 		\notag
\end{align}
The vehicle thus cannot be available for $r_1^k$ at any other location $r' <_R r$: 
\begin{align}
g_1^{r'}(r_1^k,t,q) = 0 	\notag	\\
g_2^{r'}(r_1^k,t,q) = 0	\notag
\end{align}
Concerning $r_1 = \fst(\pi_k^w)$ the first request of any other waiting location $w \ne \w(r_1^k)$, we use the same trick as for strategy $\mathcal{R}^{q}$ in order to obtain the probabilities for each possible vehicle load $q$:
\begin{align}
g_1^w(r_1,t,q) = \begin{cases}
		p_{r_1} \sum\limits_{t' = \underline{on}(w')}^{ \overline{on}(w')} \Big[ h^{w'}(\prv(r_1),t',q) + \sum\limits_{r' \in \pi_k^{w'}} h^{r'}(\prv(r_1),t',q)  \Big] , \\
		\hphantom{0}	\hspace{12em}	\text{if } ~~ t = \max( \underline{on}(w),  t^\text{min+}_{{r_1},w} )  \\ 
		0  			\hspace{12em}	\text{otherwise}.
	\end{cases} 	\notag	
\end{align}
\begin{align}
g_2^w(r_1,t,q) = \begin{cases}
		(1 - p_{r_1}) \sum\limits_{t' = \underline{on}(w')}^{ \overline{on}(w')} \Big[ h^{w'}(\prv(r_1),t',q) + \sum\limits_{r' \in \pi_k^{w'}} h^{r'}(\prv(r_1),t',q)  \Big] , \\
		\hphantom{0}	\hspace{12em} \text{if } ~~ t = \max( \underline{on}(w),  \Gamma_{r_1} )  \\ 
		0  			\hspace{12em} \text{otherwise}.
	\end{cases} 	\notag
\end{align}
with $w' = \w(\prv(r_1))$. 
From any other request $r' <_R r$ we still have:
\begin{align}
g_1^{r'}(r_1,t,q) = 0 	\notag\\
g_2^{r'}(r_1,t,q) = 0 	\notag
\end{align}
For a request $r >_R \fst(\pi_k^{w}), w \in W^x$:
\begin{align}
g_1^v(r,t, q) = \begin{cases}
		p_{r} \cdot h^v(\prv(r),t, q)   & \text{if } ~~ t >  t^\text{min+}_{{r_1},v}   \\
		p_{r} \cdot \sum_{t' = \underline{on}(w)}^{ t^\text{min+}_{{r_1},v} } h^v(\prv(r),t' , q) ~~~ & \text{if } ~~ t =  t^\text{min+}_{{r_1},v}  \\
		0 & \text{otherwise} . 	
	\end{cases} \notag
\end{align}
\begin{align}
g_2^v(r,t, q) = \begin{cases}
		(1 - p_{r}) \cdot h^v(\prv(r),t, q)   & \text{if } ~~ t > \max( \underline{on}(w),  \Gamma_r )   \\
		(1 - p_{r}) \cdot \sum_{t' = \underline{on}(w)}^{\max( \underline{on}(w),  \Gamma_r ) }    h^v(\prv(r),t' , q) ~~ & \text{if } ~~ t = \max( \underline{on}(w),  \Gamma_r )  \\
		0 & \text{otherwise} . 	
	\end{cases} 	\notag
\end{align}
when replacing $v$ by either $w = \w(r)$ or $r' \in \pi_k^w, r' <_R r, w = \w(r)$.
\\
\\
At a waiting location $w \in W^x$:
\begin{align}
h^w(r, t, q) ~ = ~  h^w_1(r, t, q) ~ + ~ h^w_2(r, t, q) ~ + ~ h^w_3(r, t, q).	\notag
\end{align}
The aforementioned terms of the sum are:
\begin{align}
h^w_1(r, t, q) = \begin{cases}
		g_1^w(r, t^{w}, q - q_r) \cdot \delta^w(r, t^{w}, q - q_r) \\
			+ ~~ \sum_{\substack{r' \in \pi_k^w\\r' <_R r}} g_1^{r'}(r, t^{r'}, q - q_r) \cdot \delta^{r'}(r, t^{r'}, q - q_r) , &\text{if } ~~ t - d_{r,w} <  \Gamma_r^\text{next}  \\ 
		0  			&\text{otherwise}.
	\end{cases} 	\notag
\end{align}
where 
\begin{align}
\delta^v(r, t, q) &= \begin{cases}
		1, & \text{if } ~~ t   \le t^\text{max+}_{r,v} ~ \wedge ~ q + q_{r} \le Q \\
		0, & \text{otherwise.}
	\end{cases} 	\notag
\end{align}
and $t^{w}=t - d_{w, r} - s_r - d_{r, w}$, $t^{r'}=t - d_{r', r} - s_r - d_{r, w}$ and $\Gamma_r^\text{next} = \Gamma_r$ if $\nxt(r)$ exists, zero otherwise.
The second term $h^w_2$ is:
\begin{align}
h^w_2(r, t, q) = g_1^w(r, t, q) \cdot \big( 1- \delta(r, t, w) \big) + g_2^w(r, t, q).	 \notag 
\end{align}
Finally:
\begin{align}
h^w_3(r, t, q) =  
		\sum_{\substack{r' \in \pi_k^w\\r' <_R r}}
		\begin{matrix}
		& \Big[ g_1^{r'}(r, t - d_{r', w}, q) (1 - \delta^{r'}(r,  t - d_{r', w}, q) \big) \\
		& \hspace{3em} + ~~ g_2^{r'}(r, t - d_{r', w}, q) \Big] \cdot  \text{bool}(t - d_{r',w} <  \Gamma_r^\text{next} )
		\end{matrix} \notag
\end{align}
where $\text{bool}(a)$ returns 1 if the Boolean expression $a$ is true, 0 otherwise.
\\
The probability that the vehicle gets rid of request $r$ at $r$'s location is:
\begin{align}
h^r(r, t, q) ~ = ~ 	&g_1^w(r, t - d_{w, r} - s_r, q - q_r) \cdot \delta^w(r, t - d_{w, r} - s_r, q - q_r) \notag \\
			&+  \sum_{\substack{r' \in \pi_k^w\\r' <_R r}}  g_1^{r'}(r, t - d_{r', r} - s_r, q - q_r) \cdot \delta^{r'}(r, t - d_{r', r} - s_r, q - q_r)	 \notag 
\end{align}
if $t \ge \Gamma_r^\text{next}$, otherwise $h^r(r, t, q) = 0$.
\\
Finally, the probability that request gets discarded from another request $r'$ location:
\begin{align}
h^{r'}(r, t, q) =  \begin{cases}
 			g_1^{r'}(r, t ,q) \cdot (1- \delta^{r'}(r, t, q) ) ~ + ~ g_2^{r'}(r,t,q),	& \text{if } ~~ t \ge \Gamma_r^\text{next} 	\\
 			0, & \text{otherwise.}
	\end{cases} 	\notag
\end{align}

\end{document}